%% file: main.tex
\newif\ifDEBUG
\DEBUGfalse

\documentclass[10pt,twocolumn,letterpaper]{article}
\usepackage[pagenumbers]{cvpr} %

\input{preamble}

\input{typesetting/typesetting}

\definecolor{cvprblue}{rgb}{0.21,0.49,0.74}

\def\paperID{12460} %
\def\confName{CVPR}
\def\confYear{2026}

\title{Inference-Time Alignment of Diffusion Models via Evolutionary Algorithms}

\author{Purvish Jajal*\\
Purdue University\\
West Lafayette, IN, USA\\
{\tt\small pjajal@purdue.edu}
\and
Nick John Eliopoulos*\\
Purdue University\\
West Lafayette, IN, USA\\
{\tt\small neliopou@purdue.edu}
\and
Benjamin Shiue-Hal Chou\\
Purdue University\\
West Lafayette, IN, USA\\
{\tt\small chou150@purdue.edu}
\and
George K. Thiruvathukal\\
Loyola University Chicago\\
Chicago, IL, USA\\
{\tt\small gkt@cs.luc.edu}
\and
James C. Davis\\
Purdue University\\
West Lafayette, IN, USA\\
{\tt\small davisjam@purdue.edu}
\and
Yung-Hsiang Lu\\
Purdue University\\
West Lafayette, IN, USA\\
{\tt\small yunglu@purdue.edu}
}

\begin{document}
\maketitle

\begin{abstract}
Diffusion models are state-of-the-art generative models, yet their samples often fail to satisfy application objectives such as safety constraints or domain-specific validity.
Existing techniques for alignment require gradients, internal model access, or large computational budgets --- resulting in high compute demands, or lack of support for certain objectives.
In response, we introduce an inference‑time alignment framework based on evolutionary algorithms. 
We treat diffusion models as black-boxes and search their latent space to maximize alignment objectives.
Given equal or less running time, our method achieves 3-35\% higher ImageReward scores than gradient-free and gradient-based methods.
On the Open Image Preferences dataset, our methods achieve competitive results across four popular alignment objectives.
In terms of computational efficiency, we require 55\% to 76\% less GPU memory and are 72\% to 80\% faster than gradient-based methods.
\end{abstract}

\begingroup
\renewcommand\thefootnote{}\footnotetext{\textbf{*} P.~Jajal and N.J.~Eliopoulos contributed equally to this work.}
\endgroup

\section{Introduction}
\label{sec:intro}

\begin{figure}[!th]
    \centering
    \includegraphics[width=0.99\linewidth]{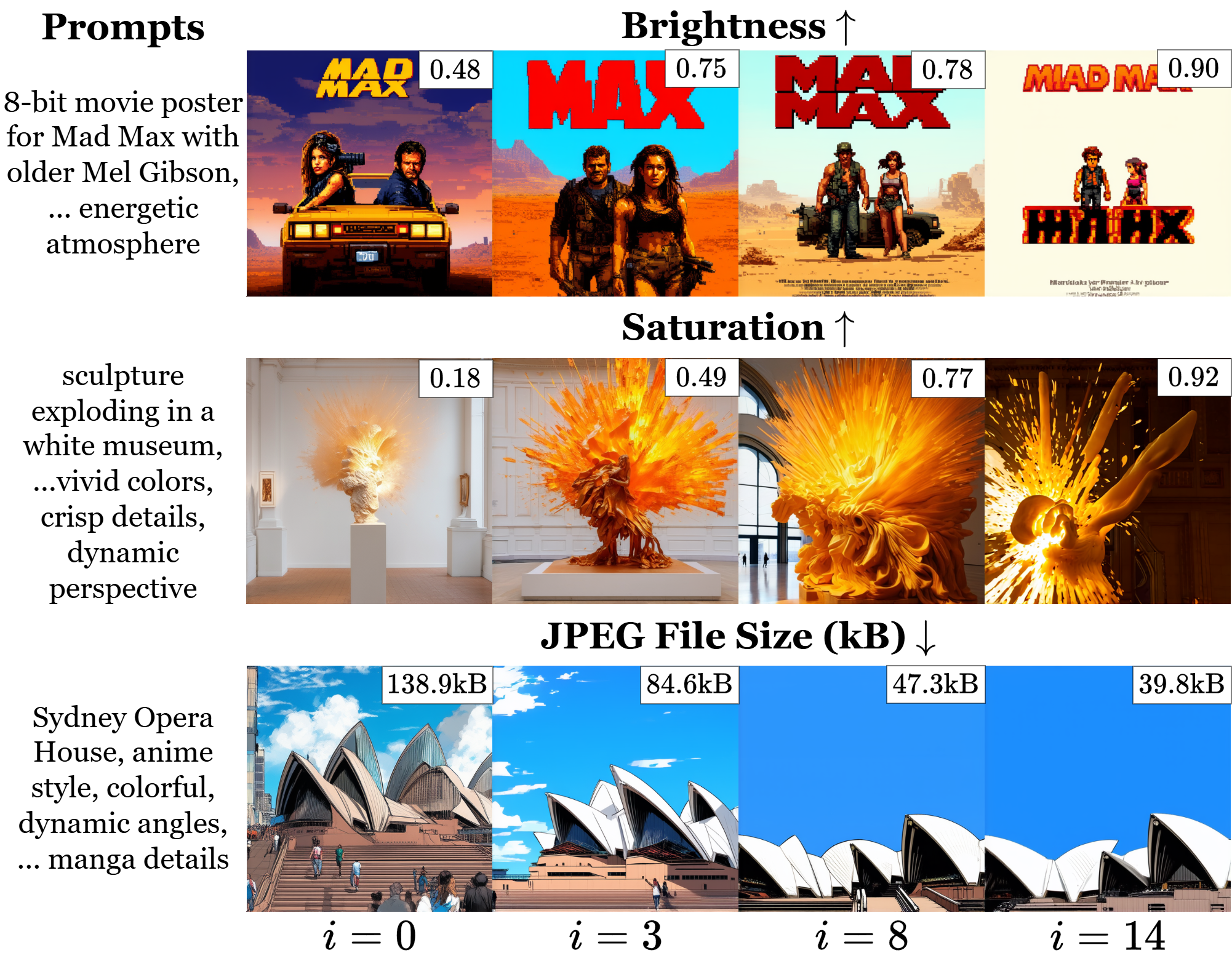}
    \caption{
    Samples generated by our method on Stable Diffusion-3.
    Each row shows the progression over optimization steps $i$,
    with the corresponding metric values displayed in the top-right corner. 
    The noise in each generation is optimized via our evolutionary-based approach, which uses the \cosyne{} algorithm~\citep{gomez2008accelerated} over 14 optimization steps.
    Arrows ($\uparrow$/$\downarrow$) indicate whether the metric is being maximized or minimized.
    }
    \label{fig:qualitative_3row}
    \vspace{-1em}
\end{figure}

Diffusion models~\citep{sohl-dickstein_deep_2015, rombach2022high, esser2024scaling} are state-of-the-art generative models for synthesizing high-quality images, video, and audio~\citep{chen2023pixart, ho2022video_diffuion, kong2020diffwave, liu2024latent}.
However, diffusion model outputs often fail to meet downstream objectives, such as user preferences~\citep{xu2023imagereward,hpsv2wu2023human,laionAestheticDataset}, safety constraints~\citep{miao2024t2vsafetybench}, or domain-specific validity~\citep{gu2024aligning, wang2024fine}.
This problem is referred to as \textit{alignment}---ensuring generated samples satisfy objectives beyond the original maximum-likelihood objective~\citep{liu2024alignmentdiffusionmodelsfundamentals, wallace2024diffusion, black2023training, lee2023aligning, dai2023emu, prabhudesai2023aligning, clark2023directly}.

We reiterate
the desirable qualities of a diffusion alignment method~\cite{uehara2025inference, tang2024inference_dno}:
  (1) \textit{black-box}, meaning it applies on a wide range of models because access to gradients or internal model states is not required;
  (2) support for \textit{arbitrary alignment objectives},
  (3) maintaining \textit{sample-efficiency}, meaning an objective can be optimized with few samples.
Most prior works capture only some of these properties~\cite{uehara2025inference}.
The works that satisfy all these properties (Best-of-N, Zero-Order) are often used as simple baselines, and we find that these baselines can be \emph{competitive} with white-box methods.
This suggests that alternate, black-box alignment formulations are worth investigating. 

In this work, we ask whether \textit{black-box evolutionary algorithms (EAs)} can be used for diffusion model alignment.
EAs should satisfy~\cite{eiben2015introduction} the aforementioned criteria, though they have not been applied to the diffusion alignment task.
We treat the diffusion model as a black box, and evolve its latent noise vector with EAs to maximize alignment objectives.
This approach enables efficient alignment supporting both differentiable and non-differentiable rewards, and is compatible with various diffusion models.
\cref{fig:qualitative_3row} depicts our method applied to various alignment objectives.

We evaluate our method in terms of its ability to achieve higher rewards than other methods under short-term optimization and equal-compute regimes.
Our method is evaluated with 
    two prompt datasets~\citep{drawbench2022,data_is_better_together_open_image_preferences_v1_flux_dev_lora_2024},
    four objective functions~\cite{xu2023imagereward,hpsv2wu2023human,clip2021,clark2023directly} (differentiable and non-differentiable),
    and on
    Stable Diffusion 1.5~\cite{rombach2022high} as in prior work.
Our method outperforms existing gradient-free and gradient-based inference-time alignment approaches on both prompt datasets.
We achieve better sample-efficiency, rewards, and memory efficiency over short optimization horizons and under equal-compute regimes.
Given similar or less running time, our method achieves 3-35\% higher ImageReward scores than gradient-free and gradient-based methods.
On the Open Image Preferences dataset, our methods achieve the highest rewards across four popular alignment objectives.
In terms of computational efficiency, we require 55\% to 76\% lower GPU memory and are 72\% to 80\% faster than gradient-based methods, while consuming up to $1.5$x more memory than gradient-free methods. 

\ul{In sum, our contributions are}:
\begin{itemize}
    \item We propose an inference-time alignment approach using evolutionary algorithms (EAs).
    Specifically, we introduce two black-box methods for aligning the outputs of diffusion models:
      (1) directly optimizing latent noise vectors,
      and
      (2) optimizing transformations applied to the noise vector.
    We elaborate on the considerations in using EAs for inference-time alignment, such as search algorithms, initialization, operators, and computational considerations.
    \item We evaluate two representative families of EAs (genetic algorithms and evolutionary strategies) for diffusion model alignment.
    Our evaluation shows generalizability over four popular reward functions, demonstrating EAs are both sample- and computationally-efficient.
\end{itemize}

\section{Background and Related Work}\label{sec:related_work}
In this section, we review prior work on diffusion model alignment (\cref{sec:related_work:alignment}) and evolutionary algorithms (\cref{sec:rw:evo_alg}).
\cref{sec:appendex:ex_diff_related} gives an extended treatment of related works.

\subsection{Diffusion Model Alignment}\label{sec:related_work:alignment}
Diffusion models use a reverse diffusion process to convert a latent noise distribution into a data distribution, such as images~\citep{sohl-dickstein_deep_2015,ho2022video_diffuion}.
The reverse diffusion process (sampling) iteratively denoises the initial latent noise $z_T$ over some number of steps $T$ 
to yield a sample, $z_0$.
This denoising is guided by conditioning on some variable, usually a text-based prompt.
Despite this, they may fail to produce samples that meet some downstream objective.
\textit{Diffusion model alignment methods} adjust diffusion models such that the resulting samples better meet an objective beyond the model's original maximum likelihood criterion. 
Examples include aesthetics~\citep{kirstain2023pick, hpsv2wu2023human, hpswu2023human, xu2023imagereward}, or compressibility~\citep{black2023training}.

Alignment methods can be grouped into two broad categories: fine-tuning-based methods and inference-time methods~\cite{uehara2025inference}.
Fine-tuning-based alignment methods involve adjusting the diffusion model's parameters so that generated samples better match alignment objectives~\citep{lee2023aligning, dai2023emu, prabhudesai2023aligning, clark2023directly, wallace2024diffusion,black2023training}.
Fine-tuning based methods, although powerful, require retraining and thus all the associated costs.
We follow the alternative approach, \emph{inference-time methods}. 
Rather than altering model parameters, these techniques adjust the diffusion sampling or conditioning to ensure samples meet alignment objectives.
Formally, they seek the control variable $\psi$ that maximizes the expected reward $R(x)$ of samples $x$ from a pretrained diffusion model $p_\theta$, as shown in~\cref{eq:unified_inftime}.

\begin{equation}
\label{eq:unified_inftime}
    \psi^\star = \underset{\psi\in\Psi}{\arg\max}\;\mathbb{E}_{x\sim p_\theta(x\mid \psi)}\bigl[R(x)\bigr]
\end{equation}

Examples of control variables $\psi$ include optimizing conditional input prompts~\citep{hao2023optimizing, gal2022image, manas2024improving, fei2023gradient}, manipulating cross-attention layers~\citep{feng2022training}, and 
tuning latent noise vectors (noise optimization)
to guide the diffusion trajectory~\citep{wallace2023end, tang2024inference_dno, li2025dynamic, uehara2025inference, zhou2024golden, ma2025inference_time_scaling}.
The most general control variable is noise optimization ($\psi = z$), which has two main categories: gradient-based and gradient-free methods.

\textit{Gradient-based} methods refine the noise iteratively by leveraging the gradient with respect to the reward. 
Recent examples include DOODL~\citep{wallace2023end} and Direct Noise Optimization (DNO)~\citep{tang2024inference_dno}.
These methods can incur large runtime and memory costs due to backpropagation, but can achieve high rewards over long optimization horizons.

\textit{Gradient-free} methods, on the other hand, explore the space of noise vectors or trajectories using search or sampling methods.
Uehara \etal~\citep{uehara2025inference} review inference-time algorithms, covering sampling-based and search-based methods.
Ma \etal~\citep{ma2025inference_time_scaling} employ three different search strategies --- Best-of-N, Zero-Order, and Search over Paths.
Li \etal propose DSearch~\citep{li2025dynamic} and SVDD~\citep{li_svdd_2024}, which are beam search algorithms.
Singhal \etal propose Fk Steering~\citep{singhal_fk_steering_2025} (FKS), a sampling-based alignment method.

DSearch, SVDD, and FKS methods are white-box, because they modify and rely on the diffusion denoising process.
Although Best-of-N and Zero-Order search are black-box, we show that evolutionary methods outperform them in alignment scores and sample-efficiency (\cref{sec:eval:cross_dataset,sec:eval:computational}).

\begin{figure*}[!tbh]
    \centering
    \includegraphics[width=0.95\linewidth]{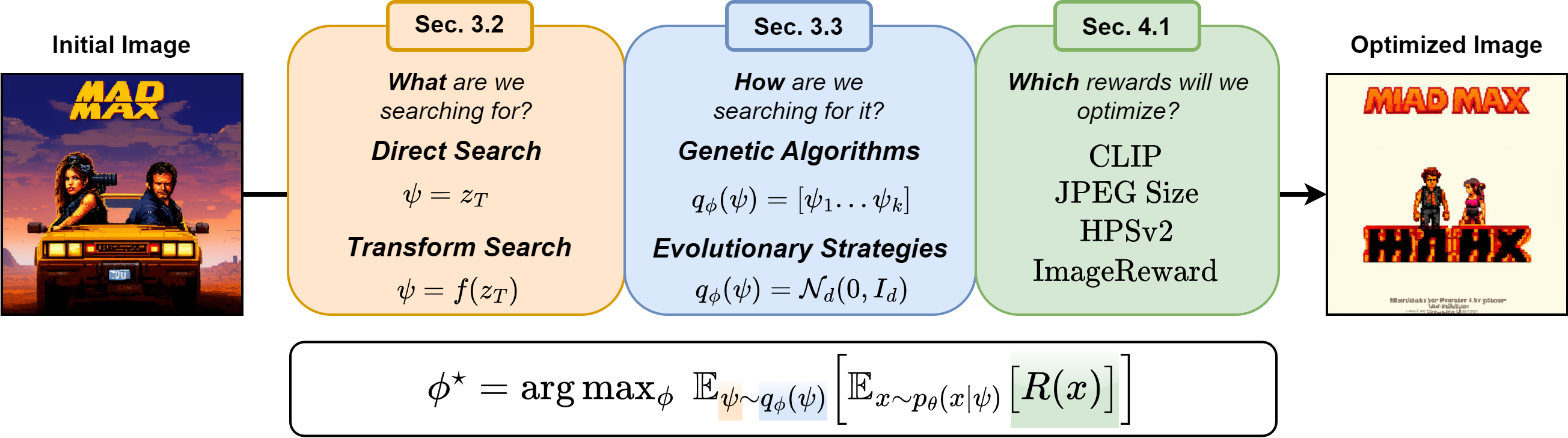}
    \caption{
    Mapping between \cref{eq:evo_general} and \cref{alg:noise_search} search over $z_T$ directly, or an affine transform of $z_T$.
    We depict connections between~\cref{eq:evo_general} and our method via color-coding.
    We perform alignment on human preferences (HPSv2, ImageReward), JPEG size, and CLIP scores.
    }
    \label{fig:direct_search_flow}
\end{figure*}

\subsection{Evolutionary Algorithms}
\label{sec:rw:evo_alg}
Evolutionary algorithms (EAs) are a class of biologically inspired methods that can be used to solve a variety of black-box optimization problems~\citep{eiben2015introduction, simon2013evolutionary}. 
They iteratively select, recombine, and mutate populations of solutions to improve their fitness (solution quality).
Among the various EAs, genetic algorithms (GAs)~\citep{eiben2015introduction,simon2013evolutionary} and natural evolutionary strategies (ES)~\citep{wierstra2014natural} are well-studied and pertinent to this work.
More details on GA and ES are in~\cref{sec:appendix:extended_commentary}.

GA and ES have been widely applied in deep learning and deep reinforcement learning.
In deep learning, genetic algorithms have facilitated neural architecture search and hyperparameter tuning~\citep{miikkulainen2024evolving, young2015optimizing}.
In deep reinforcement learning, both genetic algorithms and natural evolutionary strategies have been shown to be competitive with back-propagation-based methods such as policy gradients~\citep{such2017deep, salimans2017evolution}.
GA and ES are also used in image generation~\citep{lehman2011evolving, secretan2008picbreeder, tian2022modern, fei2023gradient} and artificial life simulations~\citep{kumar2024automating}.
We are the first to apply and parameterize EAs for black-box diffusion model alignment.

\section{Inference-Time Alignment via Evolutionary Algorithms}
\label{sec:method}

We cast inference‐time alignment as a black-box search problem and apply an evolutionary search framework.
\cref{fig:direct_search_flow} visually depicts our method, which we now explain.

Given a pretrained diffusion model $p_\theta$ and reward function $R(x)$ corresponding to some alignment objective, we aim to find parameters $\phi$ of the search distribution $q_\phi(\psi)$ that maximizes the expected reward in~\cref{eq:evo_general}.

\begin{equation}
\label{eq:evo_general}
\phi^\star
= 
\arg\max_{\phi}\;
\underbrace{\mathbb{E}_{\psi\sim q_\phi(\psi)}}_{\substack{\text{search}\\\text{distribution}}}
\Bigl[
  \mathbb{E}_{x\sim p_\theta(x\mid \psi)}
  \bigl[R(x)\bigr]
\Bigr]\,.
\end{equation}

For generality, we write $x\sim p_\theta(x\mid\psi)$ due to many diffusion samplers being stochastic.
In practice, we treat this as a deterministic map from control variable to sample, \ie~$f_\theta: \psi \mapsto x$. 
See~\cref{sec:appendix:black_box} for further details.

In~\cref{sec:method:s_noise} we define the solution spaces ($\psi$), over both noise and noise‐transformations.  
In~\cref{sec:method:evo_genetic} we describe how genetic algorithms and natural evolutionary strategies define the search distribution $q_\phi(\psi)$ and how they optimize $\phi$ in~\cref{eq:evo_general}.
\cref{sec:method:putall} discusses implementation considerations.

\subsection{Defining the Solution Space}
\label{sec:method:s_noise}
Before applying evolutionary algorithms, we must define the \textit{solution space} $\psi$, \ie what we will search over.
For the diffusion alignment problem, we explore two options: (1) directly searching over \textit{noise vectors}, and (2) searching over \textit{transformations} of an initial noise vector.
In both cases, our overarching goal is to identify a solution that maximizes~\cref{eq:evo_general}.

\mypar{Searching over Noise}
The simplest solution space we can define is over the initial noise, \ie $\psi=z_T'$ in~\cref{eq:evo_general}, as depicted in~\cref{alg:noise_search}.
Concretely, we define the search variable $\psi$ to be the initial noise vector, $z_T'\sim q_\phi(z_T')$, and optimize $\phi$ to maximize the expected reward. 
This direct search is straightforward, but hinges on careful parameterization; otherwise  solutions can drift into low-density regions of the latent space, leading to poor sample quality (\cref{sec:method:putall}).

\input{algos/noise_search}

\mypar{Searching for Noise Transformations}
Alternatively, to ensure residence in the valid latent space, we can instead search for affine transformations of the initial noise vector ($z_T$).
This requires only slight modification to \cref{alg:noise_search} (\cref{sec:appendix:noise_transform_impl}).
We define the transformed noise as $z_T^{\prime} = f(z_{T}) = A z_{T} + b$, where $A \in \mathbb{R}^{d\times d}$, and $z_T, b \in \mathbb{R}^{d}$.
Thus, in~\cref{eq:evo_general} we set $\psi=[A,b]$ with $[A, b] \sim q_\phi([A,b])$ and optimize $\phi$ to maximize the expected reward in~\cref{eq:evo_general} under $x\sim p_\theta(x\mid Az_T+b)$.

In general, $z_T'$ may not reside in the high-density shell of $\mathcal{N}(0,I)$, leading to poor sample quality. 
To keep $z_T'$ within the high-density shell, we can ensure $A$ is orthonormal.
If $A$ is orthonormal and $b=0$, $z_T'$ stays within the high-density shell of $\mathcal{N}(0,I)$ by the rotational invariance property of the Gaussian~\cite{gaussian_annulus_lecture_2017}.
In this work, we use the QR decomposition to extract the orthonormal component $Q(A)$,
and apply $Q(A)$ to $z_T$.
Importantly, we perform QR decomposition only \textit{on the channel dimension of $z_T$} 
to avoid large compute costs.

\mypar{Direct Noise vs. Noise Transform Search}
Direct noise search, while being the straightforward way to define our search space, does not confine solutions to the high-density shell of the Gaussian.
If the search algorithm fails to ensure shell-confinement, sample quality degrades.
In contrast, searching over orthonormal noise transformations will guarantee residence within the high-density shell 
given~$b=0$, as discussed above.

\subsection{Searching the Solution Space}
\label{sec:method:evo_genetic}

Given the solution space ($\psi$), we now turn to \textit{how} we define the search distribution and its parameters $q_\phi(\psi)$. 
We consider two major families of evolutionary algorithms (EAs) that have found prior use in DL: genetic algorithms (GAs) and natural evolutionary strategies (ES)~\cite{kramer2017genetic, eiben2015introduction}.
GA and ES should satisfy the criteria laid out in~\cref{sec:intro}, namely: black-box, sample-efficient, and supporting arbitrary rewards~\cite{kramer2017genetic, eiben2015introduction}. 
We first review their core mechanics, then map them onto the diffusion alignment task.
Their optimization procedures are outlined in~\cref{sec:appendix:evo_algs}.

\begin{figure}[!bt]
    \centering
    \includegraphics[width=0.95\linewidth]{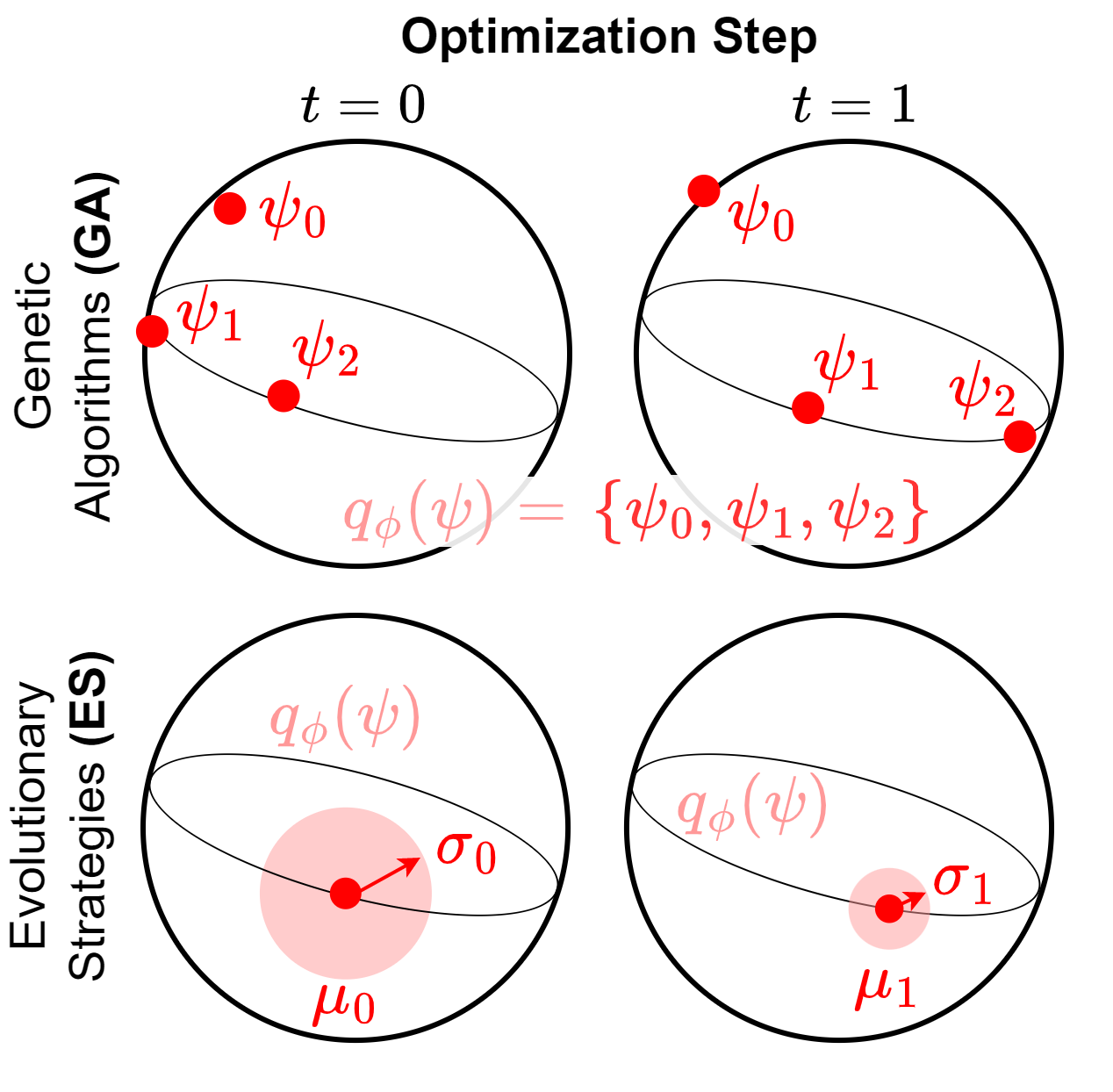}
    \caption{
    We depict how GA and ES search the latent noise space (a Gaussian hypersphere~\cite{gaussian_annulus_lecture_2017}) over optimization steps $t$.
    GAs maintain an empirical population of solutions $q_\phi=\{\psi_0...\psi_N\}$ while ES maintains a distribution $q_\phi=\mathcal{N}(\mu_i,\sigma_i)$ from which we sample.
    }
    \label{fig:sphere_search_viz}
\end{figure}

\mypar{Genetic Algorithms (GAs)}
Genetic algorithms maintain a population of candidate solutions, $\psi_i$, and iteratively improve them via selection, crossover, and mutation~\citep{eiben2015introduction}. 
In~\cref{eq:evo_general}, the search distribution is an empirical distribution over the population of solutions, as shown in~\cref{fig:sphere_search_viz}.

GAs offer practical advantages that improve alignment performance.
First, the optimization procedure---selection, crossover, mutation---under correct parameterization can ensure the noise search is regularized to the high-density shell, provided the initial population resides within. 
In particular, \textit{uniform crossover} (independent coordinate swaps) 
and \textit{permutation}-based mutations 
do so.\footnote{See our proof in~\cref{sec:appendix:gaussian_uniform} for details.}
Second, maintaining a population enables broad, parallel exploration early in optimization, and selection pressure later (\cref{sec:appendix:selection}).
This makes GAs well-suited to multimodal alignment objectives, \eg human preferences.
However, we find that GAs can suffer from low sample diversity, leading to poor alignment over long optimization horizons (\cref{sec:eval:steps_and_alignment,sec:eval:pop_stats}).

\mypar{Natural Evolutionary Strategies (ES)}  
Natural evolutionary strategies perform black-box optimization by adapting the parameterized search distribution $q_\phi$ over solutions~\citep{wierstra2014natural}. 
Unlike GAs, $q_\phi$ for ES is a parameterized search distribution (\eg multivariate normal), depicted in~\cref{fig:sphere_search_viz}.

ES offer different tradeoffs than GAs. %
They maintain a search distribution, so their memory footprint is lower: only distribution parameters (\eg mean, covariance, step size) need be stored.
This parameterized distribution also lets us draw an arbitrary number of candidate solutions, making ES ideal when many aligned samples are required.
Moreover, although ES are ``gradient free'' in the sense of no backpropagation through network parameters, they update their search distributions with a gradient approximation~\citep{salimans2017evolution, such2017deep}.
Thus, ES move through reward landscapes with a sense of direction that GAs lack, resulting in improvements over long optimization horizons (\cref{sec:eval:steps_and_alignment}).

\subsection{Implementation Considerations}
\label{sec:method:putall}
This section discusses practical considerations for implementing evolutionary algorithms for alignment.

\mypar{Initialization and Sample Quality}
Initialization is essential for effective search.
For GAs, initializing the population from the standard Gaussian distribution $\mathcal{N}(0, I)$ ensures that the initial samples reside within the high-density shell of the latent space. 
However, this strategy does not translate to ES, as subsequent updates to the search distribution can degrade sample quality.
To mitigate this, we instead initialize $\mu$ of the search distribution to a fixed latent vector $z_0 \sim \mathcal{N}(0, I)$ and use a small initial standard deviation.
\cref{sec:appendix:es_init} elaborates on the consequences of ES initialization.

\mypar{Parallel Evaluation and Population Scaling}
GA and ES approaches evaluate a set of $N$ candidate solutions at each iteration. 
To improve runtime efficiency, we partition $N$ solutions into batches of size $B$ and evaluate the batched solutions in parallel. 
This trades higher memory usage for lesser runtime costs due to inefficient serial processing. 
Detailed performance analysis is provided in~\cref{sec:eval:computational}.

\mypar{Composability}
Our evolutionary methods are compatible with other methods.
Direct or transform noise search only modifies the initial noise $z_T$.
Methods that do not target $z_T$ can be used with ours.
We demonstrate this by composing our method with a fine-tuning approach (\cref{sec:eval:finetune_compose}).

\section{Evaluation}
\label{sec:experiments}
We compare our EA alignment methods with other state-of-the-art inference-time alignment work.
\cref{sec:setup} describes our experimental setup.
\cref{sec:eval:cross_dataset} compares the ability of our method to maximize rewards with other works.
\cref{sec:eval:pop_stats} compares population statistics between ES and GA methods.
\cref{sec:eval:finetune_compose} investigates the effect of performing inference-time alignment on top of finetuning-based alignment.
\cref{sec:eval:computational} reports the memory consumption and runtime of our method across population size and batch size, including an equal-runtime evaluation.
\cref{sec:eval:steps_and_alignment} investigates the relationship between optimization steps and alignment.

\subsection{Experimental Setup}
\label{sec:setup}
We evaluate three EAs on two prompt datasets and four reward functions.
All experiments are conducted on one NVIDIA A100 80GB (PCIe) GPU.

\mypar{Algorithms}
For our approach, we use three state-of-the-art EAs: \cosyne, PGPE, and SNES (as provided in EvoTorch~\citep{toklu2023evotorch}).
In terms of state-of-the-art methods:
  we use the gradient-based approach, DNO~\citep{tang2024inference_dno};
  and three gradient-free approaches --- SVDD~\cite{li_svdd_2024},
  DSearch-R~\cite{li2025dynamic},
  and
  Fk Steering (FKS)~\cite{singhal_fk_steering_2025}.
As baseline methods,
  we follow~\citep{ma2025inference_time_scaling}
  by using
  Best-of-N search (sample $N$ images, and keep the best reward sample)
  and
  Zero-Order search (sample $N$ initial latent noises, and sample using the one with the best reward).

\mypar{Optimization Horizons}%
We evaluate our methods and DNO using optimization horizons of length 15 (``short'') or 50 (``long''), similar to DNO~\cite{tang2024inference_dno} which evaluates over horizons of length 10, 50, and 100.
We configure SVDD, DSearch-R, and FKS such that they have roughly equal runtime per prompt as our method, since they do not have a notion of optimization steps or population size in the same sense as our methods (cf.~\cref{sec:appendix:es_init:others}).

\mypar{Diffusion Models}
Here, we perform all evaluations on Stable Diffusion 1.5~\citep{rombach2022high}. 
The Appendix contains results for Stable Diffusion 3~\citep{esser2024scaling} and the latent consistency variant (LCM) of PixArt-$\alpha$~\citep{chen2023pixart}.

\mypar{Prompt Datasets}
We use two prompt datasets: \textit{DrawBench}~\citep{drawbench2022} and \textit{Open Image Preferences}~\citep{huggingface_image_preferences}.
Both datasets provide complex prompts to evaluate the quality of inference-time methods.
DrawBench comprises 200 prompts from 11 categories that test the semantic properties of model, such as their ability to handle composition, cardinality, and rare prompts.
Regarding Open Image Preferences, it features detailed prompts used to fine-tune diffusion models~\citep{data_is_better_together_open_image_preferences_v1_flux_dev_lora_2024}. 
For additional ablation evaluations, we use a randomly sampled subset (60 prompts total, 5 per category) of the full dataset. 

\mypar{Reward Functions and Measurement}
Following prior diffusion works~\citep{tang2024inference_dno, li2025dynamic},
  we experiment with four rewards for alignment:
    ImageReward~\citep{xu2023imagereward},
    CLIP~\citep{clip2021},
    HPSv2~\citep{hpsv2wu2023human},
    and
    JPEG size.
JPEG size is a proxy for compressibility, while the other three metrics evaluate image aesthetic quality for inter-method comparison.
In reporting reward statistics, we average over \textit{the dataset length, or number of prompts}. 
For granular prompt-level reward analysis,
see~\cref{sec:appendix:extended_evaluation_results}.

\input{tables/sd_drawbench_single_raw}

\input{tables/sd_opi_population_stats}

\subsection{Cross-Dataset Evaluation}\label{sec:eval:cross_dataset}
\cref{tab:eval_cross_reward} summarizes the results on DrawBench,
and
\cref{tab:opi_eval} on Open Image Preferences.
To assist interpretation, we also discuss reward hacking. %

\mypar{DrawBench}%
\cref{tab:eval_cross_reward} depicts our results on DrawBench. 
Given a short optimization horizon (15 steps), 
EAs outperform both baseline black‑box methods (Best-of-N, Zero‑Order) and the gradient‑based DNO on all four reward functions, for both the \emph{best} sample and the \emph{mean} reward in the population. 
When searching over noise, \cosyne{} delivers the strongest performance, achieving the best alignment performance on ImageReward, CLIP, and HPSv2 scores, while DSearch-R attains the lowest JPEG size.
The same pattern holds when we search over noise transformations, with \cosyne{} again leading three of the four metrics; the sole exception is the JPEG‑compression objective, where PGPE attains the smallest file sizes (however, see subsequent discussion of reward hacking). 

\mypar{Open Image Preferences}
\cref{tab:opi_eval} depicts our results on Open Image Preferences.
Similar to DrawBench (\cref{tab:eval_cross_reward}), \cosyne{} is the best performing method under equal (15) optimization steps.
\cosyne{} achieves the best alignment performance in terms of \emph{best} and \emph{mean} rewards on ImageReward and JPEG compression, and outperforming our other variants SNES and PGPE.
It achieves the highest ImageReward scores across methods, and achieve lower but similar JPEG scores to SVDD and DSearch-R.
Notably, our ImageReward scores are significantly higher than white-box methods DNO, FKS, SVDD, DSearch-R.

\mypar{Reward Hacking}
To assess the incidence of reward hacking, we audited the best method for each alignment objective, and made note of obvious cases of any reward hacking across all of its images.
On DrawBench, PGPE (ours), consistently produced images indicative of reward hacking on JPEG compression, with monochromatic results lacking prompt details.
We observed minor or one-off instances of reward hacking on other methods, which we visualize in~\cref{sec:appendix:qual:reward_hack}.

\subsection{Population Statistics}\label{sec:eval:pop_stats}

\begin{figure}[!t]
    \includegraphics[width=0.95\linewidth]{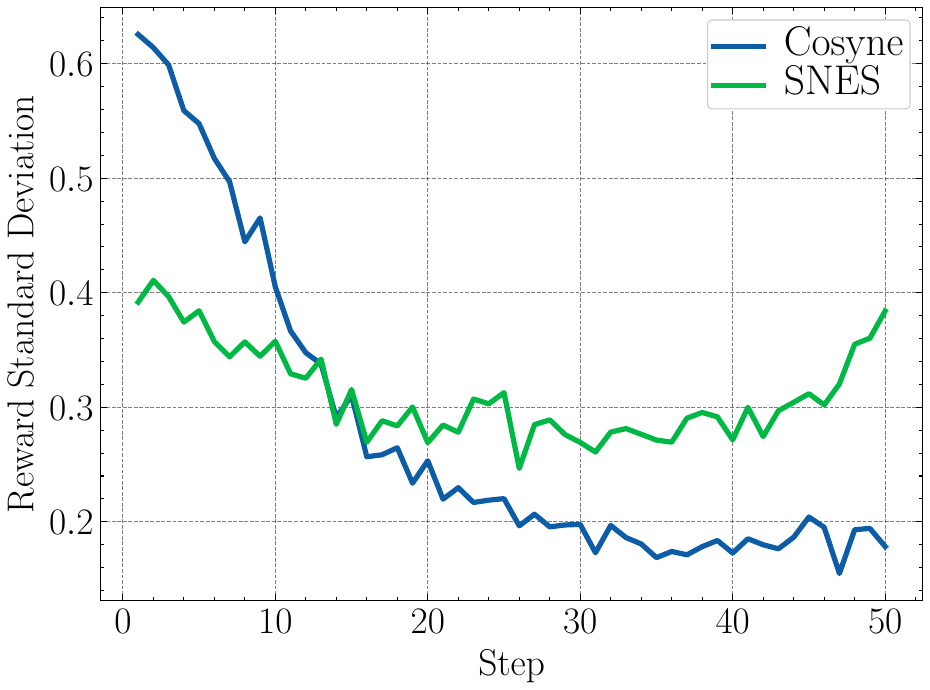}
    \caption{\textbf{Reward Standard Deviation per Step (Diversity).} 
    \cosyne{} (GA) has rapidly diminishing diversity, while
    SNES (ES) maintains diversity over optimization steps, as its search distribution evolves (\cref{sec:method:evo_genetic}).
    This implies that GAs are suited for short-term optimization, and ES for long-term optimization.
    \textit{N.B.} We use  standard deviation as a proxy for solution diversity.
    }
    \label{fig:pop_stats}
\end{figure}%
\cref{fig:pop_stats} plots the population’s reward standard deviation over 50 optimization steps for \cosyne{} (a GA) and SNES (an ES), confirming the dynamics predicted in~\cref{sec:method:evo_genetic}.
\cosyne{} begins with high variance, reflecting its broad exploration capability, but diversity collapses under selection pressure (more ablations in \cref{sec:appendix:selection}). 

\begin{figure*}[!t]
  \centering
  \begin{subfigure}[b]{0.48\linewidth}
    \centering
    \includegraphics[width=\linewidth]{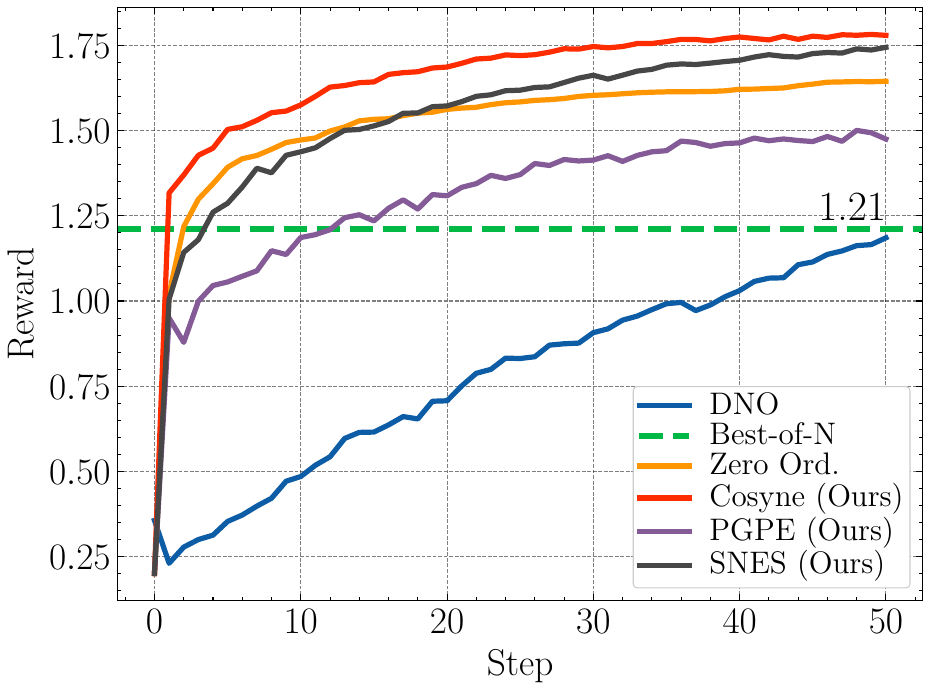}
    \caption{
    Best-sample reward per step on ImageReward.
    }
    \label{fig:ablation:pop_best_steps}
  \end{subfigure}
  \hfill
  \begin{subfigure}[b]{0.48\linewidth}
    \centering
    \includegraphics[width=\linewidth]{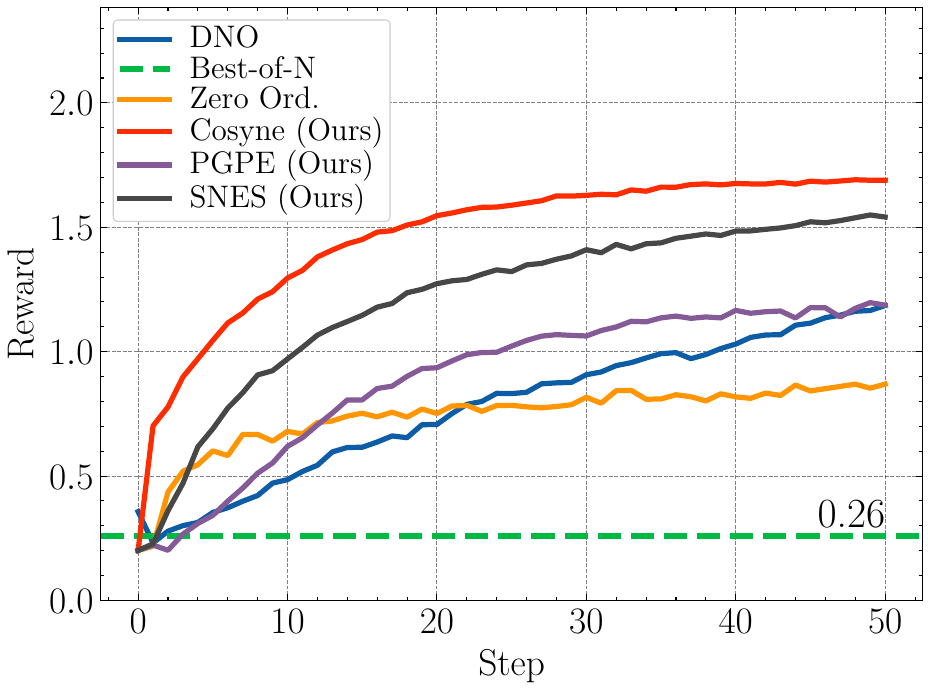}
    \caption{
    Mean reward per step on ImageReward. %
    }
    \label{fig:ablation:median_steps}
  \end{subfigure}
  \hfill
  \caption{
    Reward per-step measured on Open Image Preferences.
    Evolutionary methods (\cosyne, SNES, PGPE) outperform Zero‐Order and Best-of-N baselines. 
    Best-of-N was run with N=800.
    Algorithms were run with population size $P$=16, except DNO with $P$=1.
  }
  \label{fig:ablation:steps}
\end{figure*}

In contrast, SNES sustains a moderate level of variance throughout optimization, preserving exploratory capacity even at later steps. 
This sustained diversity underpins the long-horizon gains of ES (\cref{fig:ablation:pop_best_steps}). 
Taken together, these results suggest that GAs are ideal for rapid, short-horizon alignment; otherwise ES are preferable. %

\subsection{Composition with Fine-tuning Alignment}\label{sec:eval:finetune_compose}
\input{tables/finetuning_table}
As noted in~\cref{sec:method:putall}, our method is compatible with fine-tuning approaches.
Thus, we evaluate how our inference-time approach composes with a fine-tuning approach, Diffusion-DPO~\cite{wallace2024diffusion} on ImageReward.
\cref{tab:diff_dpo} depicts reward statistics of our approach applied to Diffusion-DPO finetuned Stable Diffusion~1.5, and the standard Stable Diffusion~1.5.
It can be seen that \cosyne{} achieves higher best and mean rewards in composition with Diffusion-DPO, indicating our method can be combined with finetuning based methods to further increase alignment.

\begin{figure*}[!htb]
  \centering
  \begin{subfigure}[b]{0.48\linewidth}
    \centering
    \includegraphics[width=\linewidth]{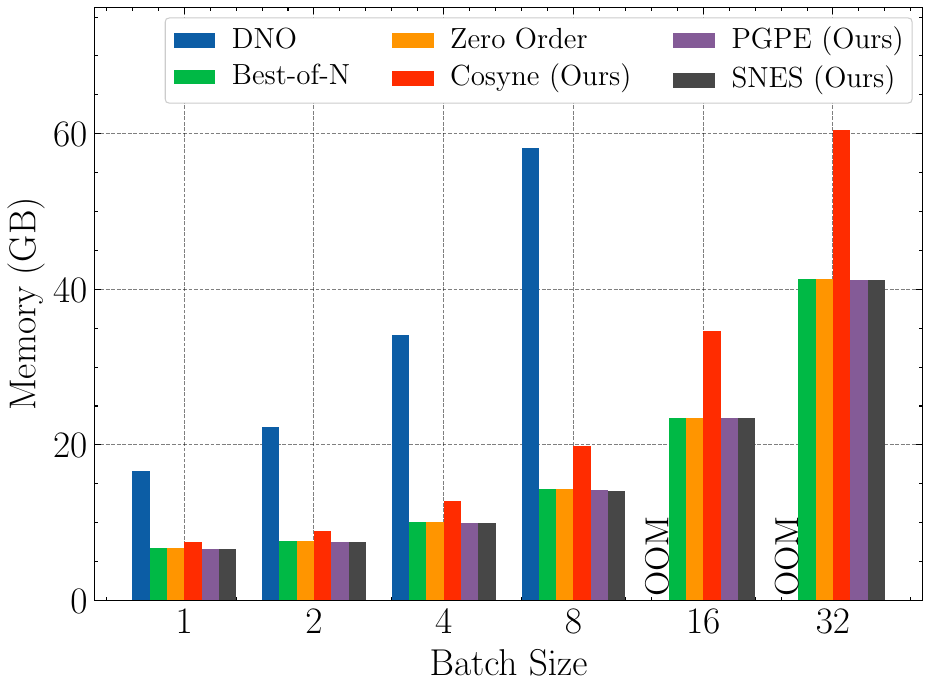}
    \caption{
    Memory usage vs. batch size ($P$=$B$) on A100 (80GB VRAM).
    }
    \label{fig:computational_costs:mem_scaling}
  \end{subfigure}
  \hfill
  \begin{subfigure}[b]{0.48\linewidth}
    \centering
    \includegraphics[width=\linewidth]{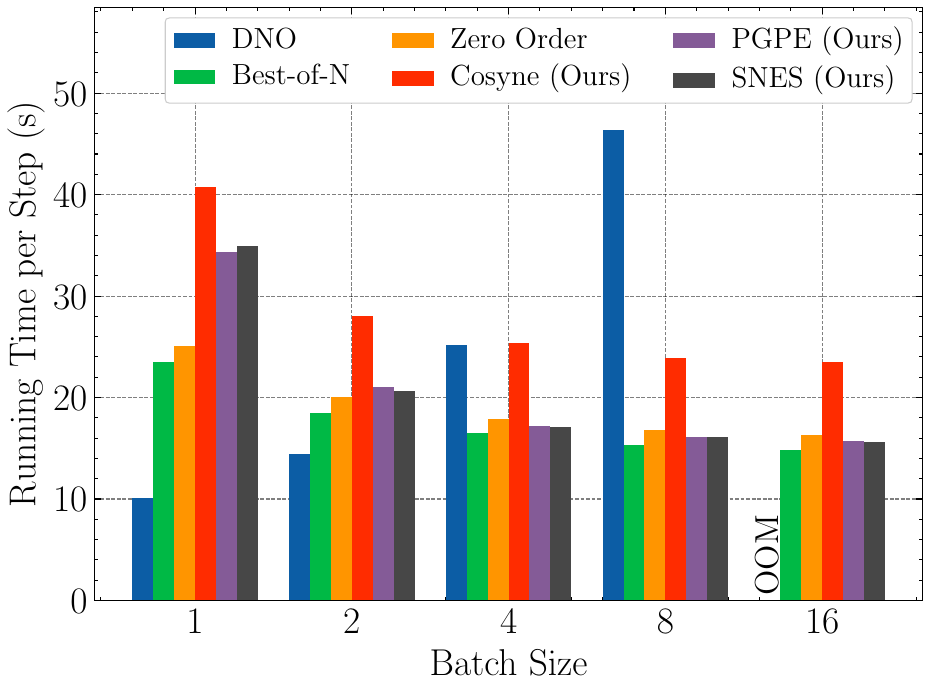}
    \caption{Running time per step vs. batch size. ($P$=16)
    }
    \label{fig:computational_costs:time_scaling}
  \end{subfigure}
  \caption{
  Computational characteristics of inference‑time alignment methods on Stable Diffusion-1.5. 
  In terms of memory usage our evolutionary methods scale with batch size, whereas DNO does not and exhausts memory for batch sizes $\geq16$. 
  Our methods exhibit near-constant or decreasing per‑step latency as batch size increases, indicating that they can be efficiently batched.
  For DNO, $P=B$.
  }
  \label{fig:computational_costs}
  \vspace{2em}
\end{figure*}

\subsection{Computational Costs}
\label{sec:eval:computational}

We discuss computational costs of surveyed methods in terms of runtime per prompt and memory consumption.

\mypar{Equal Runtime Comparison}
We depict the compute (runtime-per-prompt) required for surveyed methods in \cref{tab:equal_compute}.
In this experiment, we parameterize each method such that their runtime per-prompt nearly matches our \cosyne{} (P=16) runtime.
For the baseline Best-of-N, we assign $\sim$2$\times$ more compute to illustrate that it does not achieve high rewards, nor is it sample-efficient.
We note that \cosyne{} ($P=16$) generates an intermediate population  ($p=24$)~\cite{toklu2023evotorch,gomez2008accelerated}.
For completeness, we evaluate \cosyne{} ($P=10$) which coincides with $p=16$.

\input{tables/sd_opi_equal_compute}

\mypar{Batching Memory Comparison}
We compare the memory usage and runtime per step for evolutionary methods (\cosyne{}, PGPE, SNES) in~\cref{sec:method} with Best-of-N, Zero-Order search, and DNO.
We measure runtime per optimization step and peak GPU memory usage with respect to batch size ($B$) and population size ($P$), as stated in~\cref{sec:method:putall}.

\cref{fig:computational_costs:mem_scaling} shows that as batch size $B$ increases, evolutionary methods consume significantly less GPU memory than DNO (gradient-based).
DNO consumes $2.2\times$~to~$4.1\times$ more GPU memory than EAs, and runs out-of-memory at batch sizes $\geq16$, whereas evolutionary methods do not.
Compared with other gradient-free approaches Best-of-N and Zero-Order search, EAs consume $1\times$~to~$1.5\times$ more GPU memory, but achieve better alignment per earlier results (\cref{sec:eval:cross_dataset}).

\cref{fig:computational_costs:time_scaling} shows the running time per step for each method across batch size $B$.
All EAs exhibit a decrease in runtime per step as $B$ grows: at small $B$, the population of $P$ candidates must be evaluated sequentially, thus runtime is high.
As $B$ increases, evaluations are batched and parallelized, reducing the time per step at the expense of additional memory usage as shown in~\cref{fig:computational_costs:mem_scaling}.
In contrast, DNO's per-step runtime increases with $B$ and eventually exceeds all methods.

\subsection{Optimization Steps and Alignment}
\label{sec:eval:steps_and_alignment}
We evaluate how the number of inference‑time optimization steps affects alignment. 
\cref{fig:ablation:steps} depicts the rewards per optimization step on Open Image Preferences benchmark, using ImageReward as our alignment objective.
All methods improve their rewards as the number of steps increases, but they differ in \textit{step‑efficiency}, or reward increase per step.
\cosyne{} (a GA) achieves the highest step-efficiency, yielding the highest best-sample and mean reward at \textit{every} step count. 
Meanwhile, the EA-based methods exceed baseline mean rewards with enough steps.
PGPE needs $>$10 steps to exceed Best-of-N in best-sample performance and SNES needs $>$15 steps to beat Zero-Order search.%

\section{Limitations}
\label{sec:discussion}
Black-box methods and EAs are likely outperformed by white-box methods in terms of pure reward maximization over \emph{long optimization horizons}.
Although \cosyne{} performed better on short optimization horizons (\cref{tab:eval_cross_reward,tab:opi_eval}) than white-box approaches, its advantage diminishes in long term optimization (\cref{sec:appendix:dno_extended,fig:appendix:dno_opt_row}). 
This suggests our EA-based methods are less suitable in contexts where reward maximization is the only concern, but may be preferable when seeking high sample-efficiency.

\section{Conclusion}
\label{sec:conclusion}
We illustrate how inference-time alignment for diffusion models can be performed with evolutionary algorithms (EA).
Specifically, we search for the initial noise vector used in the reverse denoising process to maximize alignment objectives.
Our results illustrate that EAs can achieve higher rewards than existing inference-time alignment methods, particularly in short optimization horizons.
Although EAs are sample-efficient, white-box methods likely have an advantage in long optimization horizons. 
A promising direction for future work may include designing or specializing EAs to handle longer horizons.

\clearpage
{
    \small
    \bibliographystyle{ieeenat_fullname}
    \bibliography{bib/references}
}

\ifSUPPLEMENTAL
\clearpage
\input{appendix}

\fi

\end{document}

%% file: preamble.tex
\usepackage{soul}

\usepackage[utf8]{inputenc} %
\usepackage[T1]{fontenc}    %
\usepackage[pagebackref,breaklinks,colorlinks,allcolors=cvprblue]{hyperref}
\usepackage{url}            %
\usepackage{booktabs}       %
\usepackage{amsfonts}       %
\usepackage{nicefrac}       %
\usepackage{microtype}      %

\usepackage{xspace}
\usepackage{amsmath}
\usepackage{graphicx}
\usepackage{multirow}
\usepackage{wrapfig}
\usepackage{subcaption}
\usepackage{amsthm}
\usepackage{algorithm}
\usepackage{algpseudocode}
\usepackage{placeins}

\usepackage[capitalize]{cleveref}
\crefname{section}{\S}{\S\S}
\Crefname{section}{Section~\S}{Sections~\S\S}
\Crefname{table}{Table}{Tables}
\crefname{table}{Tab.}{Tabs.}

\newcommand{\ns}{\textsuperscript{\dag}} %
\newcommand{\nsp}{\textsuperscript{\phantom{\dag}}} %

\usepackage{pifont} %

\newcommand{\cmark}{\ding{51}} %
\newcommand{\xmark}{\ding{55}} %

\newcommand{\blackuparrow}{\textcolor{black}{$\uparrow$}}
\newcommand{\blackdownarrow}{\textcolor{black}{$\downarrow$}}

%% file: typesetting/typesetting.tex
\renewcommand{\cmark}{\textcolor{green!60!black}{\ding{51}}}
\renewcommand{\xmark}{\textcolor{red!70!black}{\ding{55}}}

\newcommand{\mypar}[1]{\par\textbf{#1.}~}

\newif\ifSUPPLEMENTAL
\SUPPLEMENTALtrue

\newcommand{\cosyne}{CoSyNE}

\ifDEBUG
    \newcommand{\NJE}[1]{\textcolor{red}{[NJE: #1]}}
    \newcommand{\PJ}[1]{\textcolor{blue}{[PJ: #1]}}
    \newcommand{\JD}[1]{\textcolor{purple}{[JD: #1]}}
    \newcommand{\YHL}[1]{\textcolor{olive}{[YHL: #1]}}
    \newcommand{\GKT}[1]{\textcolor{violet}{[GKT: #1]}}
    \newcommand{\GL}[1]{\textcolor{red}{[GL: #1]}}
    \newcommand{\BC}[1]{\textcolor{cyan}{[BC: #1]}}
\else
    \newcommand{\NJE}[1]{}
    \newcommand{\PJ}[1]{}
    \newcommand{\JD}[1]{}
    \newcommand{\YHL}[1]{}
    \newcommand{\GKT}[1]{}
    \newcommand{\GL}[1]{}
    \newcommand{\GR}[1]{}
    \newcommand{\BC}[1]{}
\fi

%% file: algos/noise_search.tex
\begin{algorithm}[!tbh]
\caption{Alignment via Direct Noise Search}
\label{alg:noise_search}
\begin{algorithmic}[1]
\Require Pretrained diffusion model $p_\theta$, reward $R(\cdot)$, iterations $T$, population size $M$
\State \textbf{Search distribution $q_\phi(\psi)$:} (GA) population $\{\psi_i\}_{i=1}^M$ or (ES) distribution $\mathcal{N}(\mu, \Sigma)$
\State \textbf{Initialization:} (GA) $\psi_i \sim \mathcal{N}(0,I)$ or (ES) $\mu = z_0 \sim \mathcal{N}(0,I)$, and $\Sigma = \sigma_0 I$ with a small $\sigma_0$
\For{$t = 1,\dots,T$}
    \State Sample candidates $\{\psi_i \sim q_\phi\}_{i=1}^M$ \Comment{initial noises $z_T'$}
    \State Generate samples $\{x_i\}_{i=1}^M$ with $x_i = f_\theta(\psi_i)$
    \State Compute rewards $\{r_i\}_{i=1}^M$ with $r_i = R(x_i)$
    \State $\phi \gets \text{EAUpdate}(\phi, \{\psi_i\}_{i=1}^M, \{r_i\}_{i=1}^M)$
\EndFor
\State \Return final $q_\phi$ or best candidate
\end{algorithmic}
\vspace{2pt}\noindent\footnotesize\emph{NB:} \texttt{EAUpdate} (line 7) depends on the specific evolutionary algorithm.
\end{algorithm}

%% file: tables/sd_drawbench_single_raw.tex
\begin{table*}[!ht]
  \centering
  \caption{
  \textbf{DrawBench Evaluation.} 
  We evaluate various algorithms on four popular alignment objectives.
  Algorithms (\cosyne{}, SNES, PGPE) are our evolutionary methods.
  \cosyne{}  achieves the highest rewards across all metrics, except one evaluation.
  We conduct search using each algorithm for a short-optimization horizon (15 steps) with a population of 16 solutions, except for DNO which has a population of 1.
  Best-of-N was conducted with N=240.
  \textit{N.B.} \blackuparrow/\blackdownarrow~notation indicates a goal of maximization or minimization, respectively.
  \textcolor{red!75}{Red} indicates evidence of \textcolor{red!75}{reward hacking}.
  }
  \label{tab:eval_cross_reward}
  \setlength{\tabcolsep}{5pt}
  \begin{tabular}{l  rr  rr  rr  rr}
    \toprule
    \multirow{2}{*}{\textbf{Algorithm}}
      & \multicolumn{2}{c}{(\blackuparrow) \textbf{ImageReward} }
      & \multicolumn{2}{c}{(\blackuparrow) \textbf{CLIP} }
      & \multicolumn{2}{c}{(\blackuparrow) \textbf{HPSv2} }
      & \multicolumn{2}{c}{(\blackdownarrow) \textbf{JPEG Size (kB)} } \\
    \cmidrule(lr){2-3}\cmidrule(lr){4-5}\cmidrule(lr){6-7}\cmidrule(lr){8-9}
      & Best Sample & Mean
      & Best Sample & Mean
      & Best Sample & Mean
      & Best Sample & Mean \\
      ~ & Mean $\pm$ Std & Sample & \textit{Mean $\pm$ Std} & Sample & \textit{Mean $\pm$ Std} & Sample & \textit{Mean $\pm$ Std} & Sample \\
    \midrule

    Best-of-N & $1.41 \pm 0.52$\nsp & $0.38$\nsp & $37.1 \pm 3.7$\nsp & $32.9$\nsp  & $0.301 \pm 0.02$\nsp & $0.282$\nsp & $66.9 \pm 17.8$\nsp & $105.4$\nsp \\
    Zero‑Order & $1.46 \pm 0.53$\nsp & $0.97$\nsp  & $37.6 \pm 3.9$\nsp & $34.5$\nsp  & $0.304 \pm 0.02$\nsp & $0.290$\nsp & $53.3 \pm 18.4$\nsp & $62.8$\nsp \\
    DNO & $0.71 \pm 0.93$\nsp & -\nsp & $26.2 \pm 3.6$\nsp & -\nsp & $0.286 \pm 0.02$\nsp & -\nsp & $58.1 \pm 17.5$\nsp & -\nsp \\
    FKS & $1.46 \pm 0.51$\nsp & $1.25$\nsp & $28.7 \pm 3.6$\nsp & $32.2$\nsp & $0.284 \pm 0.02$\nsp & $0.279$\nsp & $62.6 \pm 19.3$\nsp & $77.4$\nsp \\ %
    SVDD & $1.04 \pm 0.74$\nsp & $0.26$\nsp & $36.4 \pm 3.9$\nsp & $33.6$\nsp & $0.292 \pm 0.02$\nsp & $0.281$\nsp & $40.3 \pm 16.9$\nsp & $54.1 $\nsp \\ %
    DSearch-R & $0.86 \pm 0.89$\nsp & $0.72$\nsp & $36.2 \pm 4.2$\nsp & $35.5$\nsp & $0.289 \pm 0.02$\nsp & $0.286$\nsp & ${37.8 \pm 16.1}$\nsp & ${38.1}$\nsp \\ %
    \midrule
    \multicolumn{9}{c}{\textit{Ours: Direct Noise Search}}\\
    \midrule
    \cosyne      & \textbf{1.61 $\pm$ 0.42}\nsp & \textbf{1.38}\nsp  & \textbf{38.8 $\pm$ 3.8}\nsp & \textbf{36.5}\nsp  & \textbf{0.310 $\pm$ 0.02}\nsp & \textbf{0.300}\nsp  & ${39.5 \pm 15.9}$\nsp & ${44.8}$\nsp \\
    SNES        & $1.39 \pm 0.52$\nsp & $0.78$\nsp  & $38.1 \pm 3.8
    $\nsp & $35.6$\nsp  & $0.300 \pm 0.02$\nsp & $0.286$\nsp  & $71.3 \pm 21.4$\nsp & $78.0$\nsp \\
    PGPE        & $1.26 \pm 0.65$\nsp & $0.92$\nsp  & $37.1 \pm 3.6$\nsp & $34.4$\nsp  & $0.299 \pm 0.02$\nsp & $0.290$\nsp  & $88.3 \pm 23.7$\nsp & $97.0$\nsp \\
    \midrule
    \multicolumn{9}{c}{\textit{Ours: Noise Transform Search}}\\
    \midrule
    \cosyne      & ${1.53 \pm 0.44} $\nsp & ${1.23}$\nsp  & ${38.4 \pm 3.8}$\nsp & ${36.5}$\nsp   & ${0.307 \pm 0.02}$\nsp & ${0.299}$\nsp  & $38.7 \pm 16.3$\nsp & $42.3$\nsp \\
    SNES        & $1.34 \pm 0.56$\nsp & $0.94$\nsp  & $37.4 \pm 3.8$\nsp & $34.7$\nsp   & $0.300 \pm 0.02$\nsp & $0.290$\nsp  & $57.8 \pm 22.4$\nsp & $66.1$\nsp \\
    PGPE        & $1.28 \pm 0.58$\nsp & $0.88$\nsp  & $36.9 \pm 3.8$\nsp & $34.8$\nsp    & $0.300 \pm 0.02$\nsp & $0.290$\nsp  & \textcolor{red!75}{\textbf{18.3 $\pm$ 12.4}}\nsp & \textcolor{red!75}{\textbf{20.2}}\nsp \\
    \bottomrule
  \end{tabular}
\end{table*}

%% file: tables/sd_opi_population_stats.tex
\begin{table}[!t]
\centering
\caption{
\textbf{Open Image Preferences Evaluation.} 
We evaluate various methods on ImageReward and JPEG compression, configuring in the same manner as in~\cref{tab:eval_cross_reward}.
Our method based on \cosyne{} achieves the highest rewards on ImageReward, while achieving the third-lowest JPEG size.
}
\label{tab:opi_eval}
\setlength{\tabcolsep}{2pt}
\begin{tabular}{l rr rr}
    \toprule
    \multirow{2}{*}{\textbf{Algorithm}} & \multicolumn{2}{c}{(\blackuparrow) \textbf{ImageReward}} & \multicolumn{2}{c}{(\blackdownarrow) \textbf{JPEG Size (kB)}} \\
    \cmidrule(lr){2-3}\cmidrule(lr){4-5} & Best Sample & Mean & Best Sample & Mean \\
    ~ & \textit{Mean $\pm$ Std} & Sample &\textit{Mean $\pm$ Std} & Sample \\
    \midrule
    Best-of-N & 1.49 $\pm$ 0.37 & 0.49 & 72.6 $\pm$ 23.5 & 116.2 \\
    Zero–Order & 1.53 $\pm$ 0.35 & 0.98 & 55.3 $\pm$ 21.3 & 66.7 \\
    DNO & 0.78 $\pm$ 0.76 & - & 61.2 $\pm$ 17.4 & - \\
    FKS & 1.54 $\pm$ 0.38 & 1.35 & 64.5 $\pm$ 19.4 & 79.6 \\
    SVDD & 1.06 $\pm$ 0.64 & 0.46 & 44.1 $\pm$ 18.2 & 59.2 \\ %
    DSearch-R & 1.01 $\pm$ 0.66 & 0.85 &{\textbf{40.9 $\pm$ 17.7}} & {\textbf{41.3}} \\ %
    \midrule
    \multicolumn{5}{c}{\textit{Ours: Direct Noise Search}}\\
    \midrule
    \cosyne & \textbf{1.71 $\pm$ 0.24} & \textbf{1.47} & 44.2 $\pm$ 14.6 & 49.7 \\
    SNES & 1.56 $\pm$ 0.37 & 1.22 & 83.3 $\pm$ 26.1 & 90.8 \\
    PGPE & 1.42 $\pm$ 0.40 & 0.94 & 98.1 $\pm$ 34.7 & 110.3 \\
    \bottomrule
\end{tabular}
\end{table}

%% file: tables/finetuning_table.tex
\begin{table}[!tb]
    \centering
    \caption{
        \textbf{Inference-Time Alignment and Finetuning-Based Alignment.}
        We apply our inference-time alignment method on a fine-tuned model, Diffusion-DPO using ImageReward on the Open Image Preferences dataset.
        Our method improves rewards when used \emph{on top of} this fine-tuning alignment method.
        \textit{cf:}~\cref{tab:opi_eval}.
    }
    \begin{tabular}{llrr}
        \toprule
        \textbf{Model} & \textbf{Inf. Time} & \textbf{Best Sample} & \textbf{Mean} \\
        ~ & \textbf{Algorithm} & \textit{\textbf{Mean $\pm$ Std}} & \textbf{Sample} \\
        \midrule
        SD1.5 & Best-of-N & 1.49 $\pm$ 0.37 & 0.49 \\
        Diffusion-DPO & Best-of-N & 1.54 $\pm$ 0.34 & 0.66 \\
        \midrule
        SD1.5 & \cosyne & 1.71 $\pm$ 0.24 & 1.47 \\
        SD1.5 & SNES & 1.56 $\pm$ 0.37 &  1.27 \\
        Diffusion-DPO & \cosyne & \textbf{1.74 $\pm$ 0.22} & \textbf{1.55} \\
        Diffusion-DPO & SNES & 1.61 $\pm$ 0.32 & 1.31 \\
        \bottomrule
    \end{tabular}
    \label{tab:diff_dpo}
\end{table}

%% file: tables/sd_opi_equal_compute.tex
\begin{table}[!tb]
\centering
\caption{\textbf{Runtime Evaluation with ImageReward.} We use runtime as a proxy for compute, evaluating each method on Open Image Preferences with ImageReward over 50 optimization steps.
\cosyne{} achieves the highest reward for similar, and in some cases significantly more, runtime across all methods and baselines.
Methods are configured to closely match our runtime, except Best-of-N (N=2500) to exemplify its sample inefficiency.
}
\label{tab:equal_compute}
\begin{tabular}{lrr}
\toprule
\textbf{Algorithm} & \textbf{Best Sample} & \textbf{Runtime} \\
~ & \textbf{Mean} & \textbf{Per Prompt (s)} \\
\midrule
Best-of-N & 1.37 & 2104 \\ %
Zero-order & 1.66 & 1343 \\ %
DNO & 1.26 & 1270 \\ %
FKS & 1.56 & 1489 \\ %
SVDD & 1.72 & 1180 \\ %
DSearch-R & 1.15 & 1009 \\ %
\midrule
\cosyne{} ($P$=16) & \textbf{1.78} & 1119 \\ %
\cosyne{} ($P$=10) & 1.57 & 814 \\ %
SNES ($P$=16) & 1.74 & 1485 \\ %
PGPE ($P$=24) & 1.68 & 1126 \\ %
\bottomrule
\end{tabular}
\end{table}

%% file: appendix.tex
\appendix

\section{Overview of Appendices and Supplemental Material}
\label{sec:appendix:table_of_contents}
\begin{itemize}
    \item \cref{sec:appendix:table_of_contents}: This summary.\\
    \item \cref{sec:appendix:extended_commentary}: Extended commentary on related work~\cref{sec:related_work:alignment} and~\cref{sec:rw:evo_alg}, with insights and details on our implementation of evolutionary algorithms for alignment.\\
    \item \cref{sec:appendix:es_init}: Summary of insights we encountered when deciding how to initialize genetic algorithms and evolutionary strategies, related to~\cref{sec:method:evo_genetic}. We include configuration details of compared methods such as SVDD, DSearch, and FkD for fair comparison.\\
    \item \cref{sec:appendix:extended_evaluation_results}: Extended quantitative and qualitative results across various models, reward functions, datasets, and algorithms used in our work, complementary to~\cref{sec:eval:cross_dataset}---\cref{sec:eval:pop_stats}.\\
\end{itemize}

\section{Extended Commentary on Related Work and Implementations}
\label{sec:appendix:extended_commentary}
Here, we provide additional detail for concepts in our work that was not able to be fit into the main text.
We include a small table of qualitative differencesies between alignment methods (~\cref{tab:qual_alignment_latent}), which complements ~\cref{sec:related_work:alignment}.

\subsection{Extended Diffusion Model Alignment Related Work}
\label{sec:appendex:ex_diff_related}
Diffusion models use a reverse diffusion process to convert some latent noise distribution into a data distribution, such as images~\citep{sohl-dickstein_deep_2015,ho2022video_diffuion}.
The reverse diffusion process iteratively denoises the initial latent noise $z_T$ over some number of steps ($t=T \xrightarrow{} t=0$) 
to yield a sample, $z_0$.

Though diffusion models are capable of modeling complex data distributions, they often fail to produce samples that meet some downstream objective. 
\textit{Diffusion model alignment methods} adjust diffusion models such that the resulting samples better meet an objective beyond the model's original maximum likelihood criterion. 
These objectives commonly focus on producing images that reflect human aesthetic preferences~\citep{kirstain2023pick, hpsv2wu2023human, hpswu2023human, xu2023imagereward}, or other metrics such as compressibility~\citep{black2023training}. 

\input{tables/qualitative_table}

Alignment methods can generally be grouped into two main categories: fine-tuning–based methods and inference-time methods.
Fine-tuning-based alignment methods involve adjusting the diffusion model's parameters so that generated samples better match alignment objectives.
Generally, these method rely on curated datasets~\citep{laionAestheticDataset} or reward models~\citep{xu2023imagereward, hpswu2023human, hpsv2wu2023human} to fine-tune diffusion models~\citep{lee2023aligning, dai2023emu, prabhudesai2023aligning, clark2023directly}, and can involve supervised fine-tuning, or reinforcement learning based methods~\citep{wallace2024diffusion, black2023training}.
Fine-tuning based methods, although powerful, require retraining and thus all the associated costs --- our work does not require training.

Our work follows the alternative approach, \textbf{inference-time methods}. 
Rather than altering model parameters, these techniques adjust sampling or conditioning to ensure samples meet alignment objectives.
Formally, they seek the control variable $\psi$ that maximizes the expected reward $R(x)$ of samples $x$ from a pretrained diffusion model $p_\theta$, as shown in~\cref{eq:unified_inftime}.
\begin{equation}
    \psi^\star = \underset{\psi\in\Psi}{\arg\max}\;\mathbb{E}_{x\sim p_\theta(x\mid \psi)}\bigl[R(x)\bigr].
\end{equation}
Examples of control variables, $\psi$, include optimizing conditional input prompts~\citep{hao2023optimizing, gal2022image, manas2024improving, fei2023gradient}, manipulating cross-attention layers~\citep{feng2022training}, and 
tuning latent noise vectors (noise optimization)
to guide the diffusion trajectory~\citep{wallace2023end, tang2024inference_dno, li2025dynamic, uehara2025inference, zhou2024golden, ma2025inference_time_scaling}.
In this work, we focus exclusively on noise optimization ($\psi = z$), which is generalizable across diffusion models.
Broadly, noise optimization methods fall into \textit{gradient-based} optimization, or \textit{gradient-free} optimization.
We discuss each approach in turn.

\textit{Gradient-based} methods refine the noise iteratively by leveraging the gradient with respect to the reward. 
Direct Optimization of Diffusion Latents, DOODL~\citep{wallace2023end}, and Direct Noise Optimization, DNO~\citep{tang2024inference_dno}, are exemplary of these approaches.
Both need to contend with the computational costs of backpropagation and maintaining sample quality by keeping the optimized noise on the Gaussian shell.
DOODL optimizes the diffusion noise by computing the gradients with respect to a differentiable loss on generated images. 
They keep memory costs constant by leveraging invertible networks, and ensure sample quality by normalizing the optimized noise to have the norm of the original.
Likewise, DNO computes the gradient through the reward function to optimize the noise, but can optimize \textit{non-differentiable} rewards by using zero-th order optimization algorithms.  
The drawbacks of gradient-based methods are: they are not black-box; they are computationally expensive, especially when aligning multiple samples; and often require long optimization budgets (\eg DNO requires $>100$ optimization steps, whereas our method achieves better results in $\sim$50 steps). 

\textit{Gradient-free} methods, on the other hand, explore the space of noise vectors or trajectories using only search or sampling methods.
Ma \etal~\citep{ma2025inference_time_scaling}, employ three different strategies --- random search, zero-order search (similar to hill climbing), and search-over-paths --- to find samples that maximize their reward functions.
Uehara \etal~\citep{uehara2025inference} provide a comprehensive overview of various inference-time algorithms, covering sequential Monte Carlo (SMC)-based guidance, value-based importance sampling, tree search, and classifier guidance.
Li \etal propose DSearch~\citep{li2025dynamic}, a dynamic beam search algorithm in order to search for noise in order to maximize some reward function.
These techniques vary in their trade-offs, for example 
DSearch and the techniques outlined by Uehara \etal are not black-box and may have large computational costs.
Meanwhile, although random and zero-order search are black-box and computationally efficient we show evolutionary methods outperform them (\cref{sec:eval:cross_dataset}).

In this work, we investigate a different class of gradient-free optimization methods: evolutionary algorithms. 
We pursue evolutionary algorithms due to their: 
(1) ability to be used for black-box optimization; 
(2) overcoming certain computational and hardware limitations associated with existing gradient-based and complex search-based methods; 
(3) potential to effectively explore multi-modal search spaces.
As such, despite being a black-box method, we outperform both classes of methods, while remaining computationally efficient.

\subsection{Extended Alignment Objective Commentary}
\label{sec:appendix:black_box}
Here, we make additional comments upon \cref{eq:evo_general} as presented in \cref{sec:method}.
Recall that \cref{eq:evo_general}:
\begin{equation*}
\phi^\star
= 
\arg\max_{\phi}\;
\underbrace{\mathbb{E}_{\psi\sim q_\phi(\psi)}}_{\substack{\text{search}\\\text{distribution}}}
\Bigl[
  \mathbb{E}_{x\sim p_\theta(x\mid \psi)}
  \bigl[R(x)\bigr]
\Bigr]\,,
\end{equation*}
cast inference-time alignment as search problem where we find the search distribution $q_\phi(\psi)$, that maximizes the expected reward.
This objective is a generalization of \cref{eq:unified_inftime} to natural evolutionary strategies~\citep{wierstra2014natural}.
Namely, we rather than searching for a single $\psi$ that maximizes the expected reward, we assume that $\psi$ is sampled from a parameterized search distribution, $q_\phi$, and we maximize the expected reward under this distribution.
Note: $x$ is either sampled from the solution space $\psi$ in a stochastic manner
or computed deterministically. 
In this context, $\theta$ is the parameterization of the sampler of the diffusion model, and the sampler is modeled either as $x \sim p_\theta(x | \psi)$ (stochastic) or $x = f_\theta(\psi)$ (deterministic).

\subsection{Extended Characterization of Evolutionary Algorithm Components}
\label{sec:appendix:evo_algs}
Here, we provide an extended and in-depth overview of the mechanisms of genetic algorithms and natural evolutionary strategies.

\paragraph{Genetic Algorithms.}
Genetic algorithms maintain a population of candidate solutions, $\psi_i$, and iteratively improve them via selection, crossover, and mutation~\citep{eiben2015introduction}. 
In the context of~\cref{eq:evo_general}, the population is viewed as an empirical distribution. 
Thus the objective is to maximize the expected reward of a population of solutions.
This objective is maximized iteratively and at each generation:
\begin{enumerate}
  \item \textbf{Evaluation:} For each solution $\psi_i$ in the population, the fitness (reward) is evaluated --- \ie $R(x_i)$ where $x_i \sim p_\theta(x| \psi_i)$.
  \item \textbf{Selection:} Parent vectors, $\psi_i$ are chosen based on fitness, e.g.\ via tournament or fitness-proportionate selection.
  In the context of alignment,  selection will result in choosing better aligned solutions to use as parents for the next generation of solutions.
  \item \textbf{Crossover:} Pairs of parents exchange noise component (or transformation parameters) to produce offspring. 
  Correct choice of crossover can help ensure we meet (\textbf{C1}).
  For example,
  when using \emph{uniform crossover} (swapping each coordinate independently), each child’s coordinate is drawn from one of two i.i.d. $\mathcal{N}(0,1)$ parents thus remaining with high-density shell (\cref{sec:appendix:gaussian_uniform}).
  \item \textbf{Mutation:} Offspring are perturbed with noise, often in an additive manner. E.g. $x = x + \epsilon,\;\text{where}\; \epsilon \sim \mathcal{N}(0, \sigma)$. $\sigma$ is a hyperparameter of the genetic algorithm.
\end{enumerate}

\paragraph{Natural Evolutionary Strategies.}
Natural evolutionary strategies perform black-box optimization by adapting the parameterized search distribution $q_\phi$ over solutions~\citep{wierstra2014natural}. 
Unlike GAs, $q_\phi$ is a parameterized search distribution (\eg multivariate normal) and at each iteration its parameters are updated as follows: 
\begin{enumerate}
  \item \textbf{Sampling:} A population of solutions is sampled for the search distribution $\psi_i \sim q_\phi(\psi)$.  
  \item \textbf{Evaluation:}
  For each solution $\psi_i$ in the population, we evaluate the fitness as $R(x_i)$, where $x_i \sim p_\theta(x| \psi_i)$.
  \item \textbf{Update:} Estimate the gradient $\nabla_\phi \mathbb{E}_{\psi\sim q_\phi(\psi)}\Bigl[\mathbb{E}_{x\sim p_\theta(x\mid \psi)}\bigl[R(x)\bigr]\Bigr]$ and update $\phi$ to increase expected reward along the natural gradient.
  Thus, we update the parameters of the search distribution $q_\phi$ so that future samples are better aligned.
\end{enumerate}

\subsection{Noise Transform Search Implementation}\label{sec:appendix:noise_transform_impl}
Earlier we introduced the algorithm for direct noise search (~\cref{alg:noise_search}).
We noted that the algorithm for noise noise transform search is slightly different: we include it here for clarity.

\input{algos/noise_transform}

\subsection{Proof of Gaussian Marginal Preservation under Uniform Crossover}
\label{sec:appendix:gaussian_uniform}
\paragraph{Purpose} 
In this section we show that uniform crossover maintains the solutions within the high-density Gaussian shell, which was alluded to in ~\cref{sec:method:evo_genetic}.
This is relevant to the EvoTorch implementation of the CoSyne algorithm, which uses a uniform distribution to decide crossover behavior (parameterized by a probability $p$ to perform crossover).

\paragraph{Background.} Samples from a high-dimensional Gaussian lie on a hypersphere, this phenomenon is known as the Gaussian Annulus Theorem.
Diffusion models that use Gaussian noise are (implicitly) trained under this condition~\cite{gaussian_annulus_lecture_2017}. 
Thus, to ensure valid generated images, we must enforce this condition (draw samples near the surface of the sphere or shell) when we perturb noise.
Uniform crossover is one such pertrubation, which we now investigate.

\paragraph{Lemma.}
Let $X = (X_1,\dots,X_d)$ and $Y=(Y_1,\dots,Y_d)$ be independent draws from $\mathcal{N}(0,I_d)$.  
To perform coordinate‐wise uniform crossover, for each $i$ we independently sample: 
\begin{equation*}
    B_i \sim \mathrm{Bernoulli}\bigl(p\bigr),
    \quad
    Z_i = 
    \begin{cases}
    X_i, & B_i=1,\\
    Y_i, & B_i=0.
    \end{cases}
\end{equation*}
Then each coordinate $Z_i\sim\mathcal{N}(0,1)$ and thus $Z\sim\mathcal{N}(0,I_d)$.

Fix a coordinate $i$.  For any real $z$, by the law of total probability conditioning on $B_i$:  
\begin{equation*}
    \begin{aligned}
    P(Z_i \le z) &= P(Z_i \le z \mid B_i=1)\,P(B_i=1)\\
    &+P(Z_i \le z \mid B_i=0)\,P(B_i=0)\\
    &= P(B_i=1)\,P(X_i \le z)\\
    &+ P(B_i=0)\,P(Y_i \le z) \\
    &= p \,\Phi(z) + (1-p)\,\Phi(z) \\
    &=\Phi(z)
    \end{aligned}
\end{equation*}
where $\Phi$ is the standard normal CDF and we used $X_i,Y_i\sim\mathcal{N}(0,1)$.  Thus the CDF of $Z_i$ matches that of $\mathcal{N}(0,1)$, so $Z_i\sim\mathcal{N}(0,1)$.  
Since each $B_i$ acts independently on its coordinate and all coordinates of $X$, $Y$ are independent, the $Z_i$ remain independent.  Therefore the joint distribution of $Z$ is  
\begin{equation*}    
P(Z_1 \le z_1,\dots,Z_d \le z_d)
=\prod_{i=1}^d P(Z_i \le z_i)
=\prod_{i=1}^d \Phi(z_i),
\end{equation*}
which is the CDF of $\mathcal{N}(0,I_d)$.  Equivalently, $Z\sim\mathcal{N}(0,I_d)$.

\paragraph{Implication.}
As noted in~\cref{sec:method:evo_genetic}, uniform crossover will ensure that solutions are with the high-density shell of the gaussian thus maintaining sample quality.

\section{Details on Parameterization of Evolutionary Algorithms and Compared Methods}
\label{sec:appendix:es_init}
Here, we review more details on how we initialized our EA search parameters, and how other algorithms were configured for fair comparison.

\subsection{EA Parameterization}
\label{sec:appendix:es_init:ea}
\paragraph{Genetic Algorithm Mutation Rate and Stability.}
When using GAs, the mutation operator introduces Gaussian noise to promote exploration. 
However, excessive perturbations can push solutions outside the high-density shell, leading to poor sample quality. 
To address this, we use small mutation step sizes, which help ensure that offspring remain within the high-density shell and preserve sample quality.
Typically, our mutation size is $\sigma\approx0.1$.

\begin{figure}[!tb]
    \centering
    \includegraphics[width=0.95\linewidth]{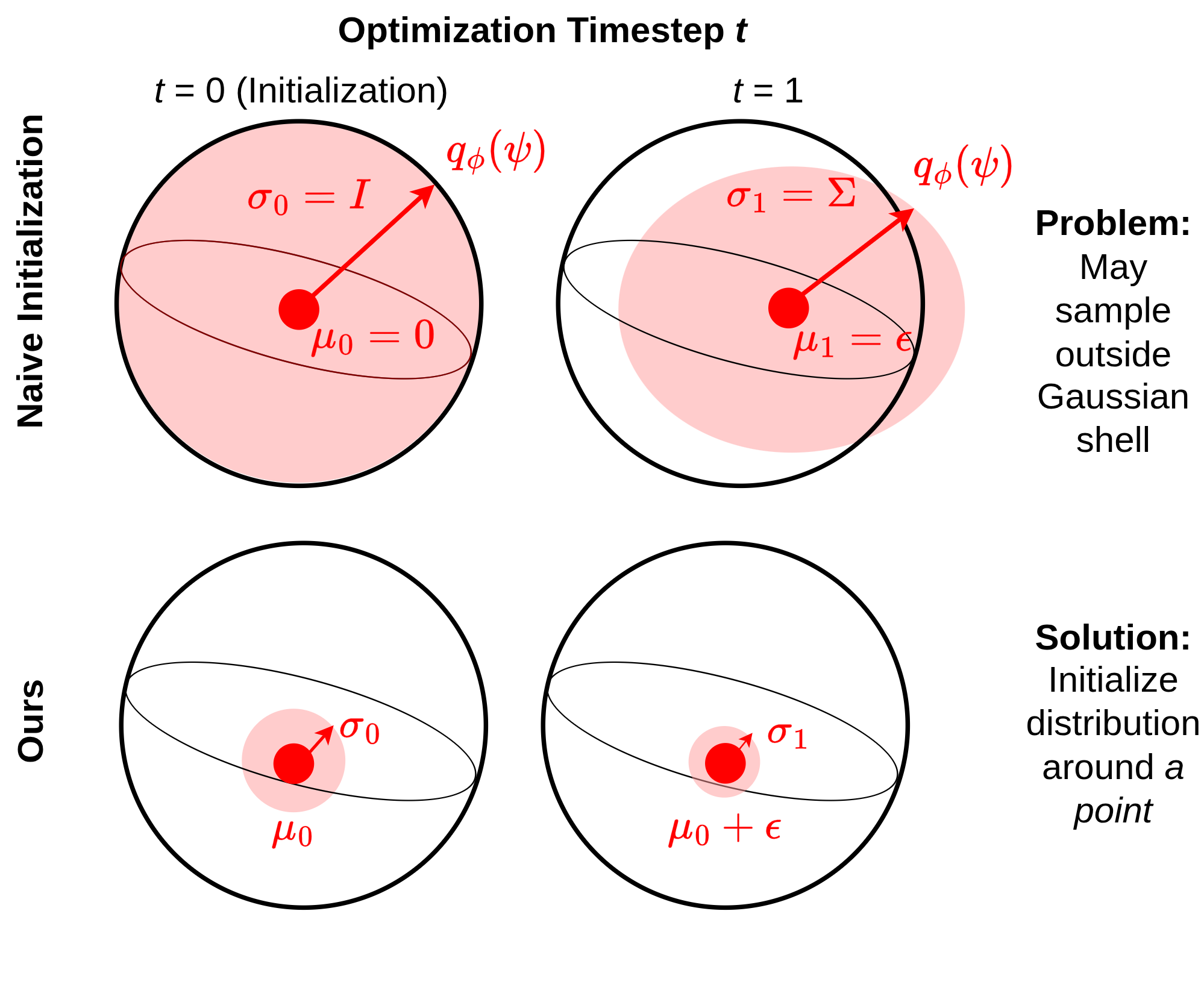}
    \caption{
    We depict a shortcoming with naive ES initialization, and illustrate our solution.
    Initializing the search distribution $q_{\phi}(\psi)$ incorrectly (zero-mean, isotropic multivariate Gaussian) can quickly lead to sampling outside the Gaussian shell, leading to poor sample quality.
    We encountered this issue in our early experiments.
    We addressed it by centering $q_{\phi}(\psi)$ on some point on the shell, and restricting its initial covariance $\sigma_0$ to a localized region, rather than encapsulating the entire shell.  
    }
    \label{fig:appendix:es_init}
\end{figure}

\paragraph{Evolutionary Strategy Initialization.}
Here, we briefly describe pitfalls with improper ES initialization, and how we addressed them in our work.
\cref{fig:appendix:es_init} visually depicts this concept.
We noted that ES are particularly sensitive to initialization --- in the context of this work, this "initialization" is the initial search distribution $q_{\phi}(\psi)$.

One naive, but simple way to initialize $q_{\phi}(\psi)$ is to model it as a zero-mean ($\mu=0$), isotropic (covariance $\sigma=I$) multivariate Gaussian distribution.
However, even after a few optimizations steps, the ES algorithm may update $\mu$ and $\sigma$ such that sampling from $q_{\phi}(\psi)$ no longer yields samples on the Gaussian shell.
As a result, sample quality is degraded significantly.
This is visually depicted in the top row of~\cref{fig:appendix:es_init}.

In order to address this problem, we instead randomly choose an initial $\mu$ that exists on the Gaussian shell, and initialize the covariance $\sigma$ to be restricted to a localized region.
This modification was effective at improving sample quality, because our samples were now more closely drawn from the surface of the Gaussian shell.

\subsection{Configuration of Compared Methods}
\label{sec:appendix:es_init:others}
\JD{This contains the right information, except where noted. It feels a bit wandering, however. Consider a reorganization somehow, but it's not a priority.}

\JD{Suggest you expand the Description column of Table 6 so that the table spans the full column. Where you write `Analogous to P', the first time you have not defined P yet! And you should explain what it means too, not just `analogous'. You could add information like `higher value increases compute cost' or similar to help the reader interpret it (shh, secretly, this is also so the table isn't silly looking)}
\NJE{Addressed feedback. Added "how" on our choice of configuration. Added bold paragraph headings to divvy it up, and updated tables.}

\mypar{Hyperparameter Description and Nuance} 
We list the primary hyperparameters of each method in \cref{tab:appendix:algorithm_param_descript}; these define the configurations reported in \cref{tab:appendix:algorithm_configs}.

Although some notions—such as “batch size” $B$—appear across multiple methods, they do not necessarily correspond to the same underlying mechanism. 
In our method, $B$ either equals the population size ($B=P$) when evaluations are fully parallelized, or is smaller than the population ($B<P$) when we process individuals in batches (\cref{sec:method:putall}). 
Other methods do not necessarily treat the batch dimension in the same manner.
Instead, their batch of particles or samples is more similar to \emph{a population of samples to generate} rather than \emph{the number of samples they evaluate per-step}.
For example, DSearch-R interprets $B$ as a number of samples to output --- but it resamples and replaces samples within the batch throughout the denoising process.

\mypar{Choosing Configurations} 
\cref{tab:appendix:algorithm_configs} shows configurations used in \cref{tab:eval_cross_reward,tab:opi_eval} . 
It was not straightforward to map our hyperparameters to those of other methods, given the differences noted above. 
To ensure a fair comparison, we therefore selected configurations that maximized alignment capability while matching the per-prompt running time of our strongest method, \cosyne{}, on each reward function.

\input{tables/method_configuration_description_table}
\input{tables/method_configuration_table}

\section{Extended Evaluation Results}
\label{sec:appendix:extended_evaluation_results}

\subsection{Cross-Model Evaluation on Open Image Preferences}
\label{sec:appendix:cross_model_opi}
\input{tables/open_img_model_ablation}

\cref{tab:eval_cross_model} analyzes three diffusion models and confirms that evolutionary algorithms largely remain competitive across models.
For both SD1.5 and SD3, the EA family consistently surpasses the black‑box baselines: \cosyne{} achieves the highest \textit{best} and \textit{median} rewards on every metric, \eg SD3‑CLIP rises from 39.4/37.4 with Zero‑Order to 40.6/39.0, while SD3‑ImageReward improves from 1.75/1.60 to 1.82/1.75. 
SNES and PGPE follow the same trend, but with smaller margins.
Generalization is weaker on PixArt‑$\alpha$, where improvements vanish, suggesting that black-box latent search may not transfer to architectures trained with alternate objectives such as LCM --- we elaborate in~\cref{sec:discussion}.
Overall, we show evolutionary search can align multiple diffusion models; however some latent spaces may be more difficult to search than others.

By virtue of our results, we show that optimizing noise with EAs is often sufficient to perform alignment (see~\cref{tab:eval_cross_reward}), though its effectiveness varies by model (\cref{tab:eval_cross_model}).
This raises two questions: 
(1) Are some latent spaces easier to search than others? 
(2) How can one design, train, or modify diffusion models to make their latent spaces easier to search?
Our reward improvements on PixArt-$\alpha$ were diminished compared to other diffusion models (\cref{tab:eval_cross_model}), suggesting some property of the latent space and/or diffusion model that affects our search efficacy.

\subsection{Effect of Longer Optimization Horizons on DNO}
\label{sec:appendix:dno_extended}
\Cref{fig:appendix:max_dno_opt,fig:appendix:mean_dno_opt,fig:appendix:median_dno_opt} depict the effect of longer optimization timelines on the best-sample, mean, and median rewards for the gradient-based method DNO, compared to gradient-free methods. 
Gradient-based require longer optimization timelines to reach the same reward as gradient-free methods.

\subsection{Population Size Ablation}
\label{sec:appendix:pop_size_ablation}
\Cref{fig:appendix:cosyne_rewards,fig:appendix:snes_rewards,fig:appendix:pgpe_rewards} depict the effect of increasing population size on the \cosyne{}, SNES, and PGPE algorithms.
All algorithms benefit from larger population sizes, however PGPE appears to benefit the most.

\subsection{Selection Pressure and Convergence}
\label{sec:appendix:selection}
\cref{fig:appendix:selection} shows the effect on increasing selection pressure (tournament size) on the median reward and standard deviation.
We note that high selection pressure (larger tournament size) results in a rapid reduction in the reward standard deviation as shown in \cref{fig:appendix:selection_std}, this is in line with prior work characterizing the effect of tournament selection~\citep{miller1995genetic}.

\input{tables/dno_long_horizon}

\input{tables/popsize_ablation}
\input{tables/ga_selection_pressure}

\subsection{Qualitative Results}
\label{sec:appendix:qual}
Here we provide qualitative and visual results.
We first address reward hacking behavior we witnessed with the PGPE algorithm optimizing for JPEG compressibility.
We then present sets of images that illustrate the low sample diversity of genetic algorithms across models and datasets.

\subsubsection{Reward Hacking in \cref{sec:eval:cross_dataset}}
\label{sec:appendix:qual:reward_hack}
\begin{figure*}
    \centering
    \includegraphics[width=0.95\linewidth]{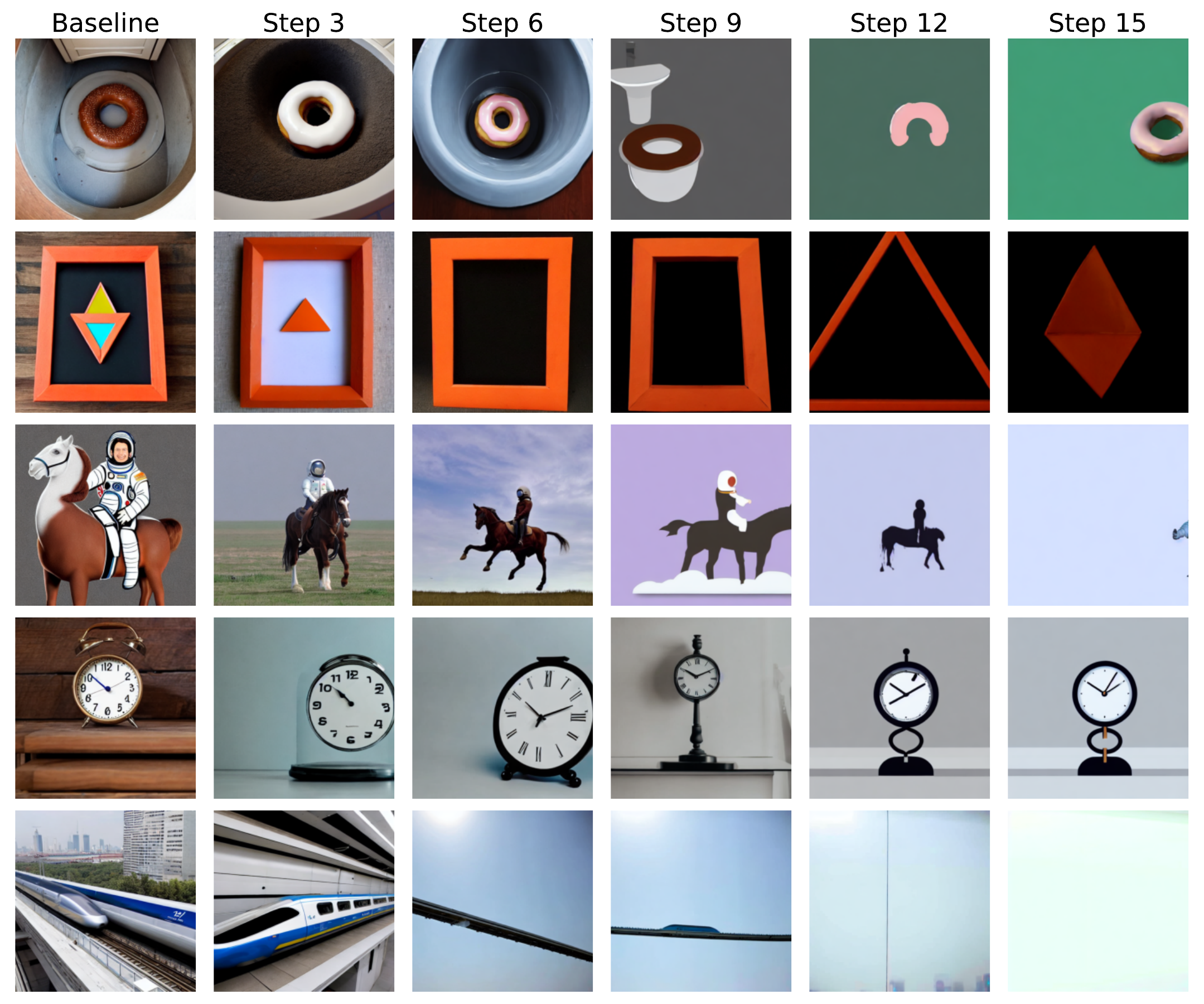}
    \caption{
    \textbf{PGPE Reward Hacking:} 
    Randomly selected images created using inference-time alignment with the PGPE algorithm (transformation search) to minimize the JPEG size (DrawBench).
    PGPE neglects the prompt, and naively attempts to minimize JPEG file size by producing monochromatic or simple images.
    }
    \label{fig:appendix:pgpe_hack}
\end{figure*}

Here we discuss reward-hacking behavior on our method, PGPE, and minor instances across other surveyed methods.
\cref{fig:appendix:pgpe_hack} shows the proclivity of PGPE to "reward-hack" as noted in \cref{sec:eval:cross_dataset}.
In many instances, PGPE was able to reduce the JPEG file size, but the resulting images were far too dissimilar or nonsensical compared with the intended image for a prompt.
This behavior \textit{is} reward-hacking, because the algorithm completely ignores key prompt details, and instead produces nonsensical images to maximize reward.

\subsubsection{Stable Diffusion Qualitative Results}
Here we illustrate the differences between ES and GA in longer optimization horizons.
We show reward statistics in \cref{fig:appendix:sd_cosyne_img_db}, \cref{fig:appendix:sd_cosyne_img_open},
\cref{fig:appendix:sd3_cosyne_img_open},
and
\cref{fig:appendix:sd3_snes_img_open}.
GAs like \cosyne{} feature low sample diversity, even as optimization steps scale.
This can also happen to ES methods (\cref{fig:appendix:sd3_snes_img_open}, however we noted it was less likely to occur with ES.
Over longer optimization horizons, ES was able to maximize rewards better than GA while having higher diversity (\cref{sec:eval:pop_stats}).

\input{tables/qualitative_diversity}

\subsection{Extended DrawBench Results}
\label{sec:appendix:ex_drawbench}
Here we include additional, granular results on the DrawBench dataset.
We provide this data in \cref{tab:appendix:imagereward_extended}, \cref{tab:appendix:hpsv2_extended}, \cref{tab:appendix:clip_extended}, and \cref{tab:appendix:jpeg_extended}.
Specifically, we include additional statistics (min, max, median, mean, standard) for each reward function.
Each subtable for a particular reward function displays the population-best rewards and median rewards, which were measured across all prompts of the DrawBench dataset.

\subsection{Total Reward Function Evaluations}
\label{sec:appendix:reward_fn_table}

The cost of evaluating the reward function is significantly higher than the cost of sampling~\cite{uehara2025inference} --- \eg human feedback.
Therefore, we report the number of \textit{total reward function evaluations} from \cref{tab:eval_cross_reward,tab:opi_eval}, which we illustrate in \cref{tab:appendix:reward_evals}.
The evaluation costs are relatively low for the alignment objectives in this work.
Even so, a method that can achieve high rewards with fewer samples is desirable in any context.

We show our methods took \emph{substantially fewer} reward evaluations than similar works FKD, SVDD, and DSearch-R, while achieving \emph{higher} aesthetic scores than all other methods.

For clarity, we describe how we derived the evaluation counts that appear in ~\cref{tab:appendix:reward_evals}, which are based on the configurations from~\cref{tab:appendix:algorithm_configs}.

\mypar{Best-of-N, Zero-Order} 
These methods were parameterized by $P=16$, run over 15 optimization steps for 240 total evaluations.
Each sample in the population was evaluated once per optimization step.

\mypar{\cosyne{}}
\cosyne{} generates an intermediate population size that is $1.5\times$ its input and output population size $P$, as mentioned in~\cref{sec:eval:computational}.
Given $P=16$, an effective population of $P=24$ is realized; over 15 optimization steps this leads to 360 total evaluations.

\mypar{FKS}
FKS resampled latents a total of 41 times (denoising step 5 until step 45) over a population of 128 particles for 5258 total evaluations.
Each particle is evaluated once at each denoising step.

\mypar{SVDD}
SVDD maintains 120 candidate latent noise vectors over 49 denoising steps for 5880 total evaluations.
Each candidate is evaluated once at each denoising step.

\mypar{DSearch-R}
DSearch-R maintains 5 candidate latent noise vectors, and follows a schedule to further expand its search, yielding $\sim$7880 evaluations under their default (exponential) search schedule.
This resamples latents at $\sim$38 denoising steps.

\input{tables/method_reward_evaluation_counts}

\FloatBarrier
\input{tables/sd_drawbench_extended}

%% file: tables/qualitative_table.tex
\begin{table}[!htb]
    \centering
    \caption{
    Qualities of diffusion alignment methods. 
    }
    \label{tab:qual_alignment_latent}
    \begin{tabular}{lcccc}
        \toprule
        \multirow{2}{*}{\textbf{Method}} & \textbf{Arbitrary} & \textbf{Gradient} & \textbf{Black} \\ %
        ~ & \textbf{Rewards?} & \textbf{Free?} & \textbf{Box?} \\
        \midrule
        DNO~\cite{tang2024inference_dno} & \cmark & \xmark & \xmark \\
        Diffusion-DPO~\cite{wallace2024diffusion} & \cmark & \xmark & \xmark \\
        DSearch-R~\cite{li2025dynamic} & \cmark & \cmark & \xmark \\
        SVDD~\cite{li2025dynamic} & \cmark & \cmark & \xmark \\
        FKS~\cite{singhal_fk_steering_2025} & \cmark & \cmark & \xmark \\
        \midrule
        Zero-Order~\cite{ma2025inference_time_scaling} & \cmark & \cmark & \cmark \\
        Best-of-N & \cmark & \cmark & \cmark \\
        \textbf{Ours} & \cmark & \cmark & \cmark \\
        \bottomrule
    \end{tabular}
\end{table}

%% file: algos/noise_transform.tex
\begin{algorithm}[!ht]
\caption{Alignment via Noise Transformation Search}
\label{alg:noise_transform}
\begin{algorithmic}[1]
\Require Pretrained diffusion model $p_\theta$, reward $R(\cdot)$, noise $z_T \sim \mathcal{N}(0,I)$, iterations $T$, population size $M$
\State \textbf{Search variable:} $\psi = A$, with transformed noise $z_T' = Q(A)\,z_T$
\Statex \quad where $Q(A)$ denotes the orthonormal component of $A$ obtained via QR decomposition
\State \textbf{Search distribution $q_\phi(\psi)$:} \; GA: population $\{A_i\}_{i=1}^M$ or (ES) parameterized $A$
\For{$t = 1,\dots,T$}
    \State Sample transformation parameters $\{A_i \sim q_\phi\}_{i=1}^M$
    \State Compute transformed noises $\{z_i' = Q(A_i)\,z_T\}_{i=1}^M$
    \State Generate samples $\{x_i\}_{i=1}^M$ with $x_i = f_\theta(z_i')$
    \State Compute rewards $\{r_i\}_{i=1}^M$ with $r_i = R(x_i)$
    \State $\phi \gets \text{EAUpdate}(\phi, \{A_i\}_{i=1}^M, \{r_i\}_{i=1}^M)$
\EndFor
\State \Return final $q_\phi$ or best transform
\end{algorithmic}
\vspace{2pt}\noindent\footnotesize\emph{NB:} \texttt{EAUpdate} depends on the specific evolutionary algorithm used. A bias term $b$ may be included, but we set it to 0.
\end{algorithm}

%% file: tables/method_configuration_description_table.tex
\begin{table*}[!thb]
    \centering
    \setlength{\tabcolsep}{24pt}
    \caption{
        \textbf{Hyperparameters of Surveyed Methods.} 
        This is a companion table for \cref{tab:appendix:algorithm_configs}.
        We refer the reader to the papers (and respective codebases) for a more detailed explanation of each.
        Batch size $B$ is interpreted differently by each method, per~\cref{sec:appendix:es_init:others}.
    } 
    \begin{tabular}{lll}
        \toprule
        Algorithm(s) & Parameter & Description \\
        \midrule
        \multirow{2}{*}{\cosyne{}, SNES, PGPE} 
            & $P$ & Population size \\
            & $B$ & Batch size, parallel evaluations \\
            & $O$ & Optimization Steps \\
        \midrule
        \multirow{4}{*}{FKS\cite{singhal_fk_steering_2025}} 
            & $\text{PC}$ & Particle Count, similar to population size $P$ \\
            & $t_{\text{start}}$ & Resampling denoising step start \\
            & $t_{\text{end}}$ & Resampling denoising step end \\
            & $t_{\text{freq}}$ & Resampling frequency \\
        \midrule
        \multirow{3}{*}{DSearch-R\cite{li2025dynamic}} 
            & $B$ & Similar to population size $P$ \\
            & $n_\text{duplicates}$ & Beam width \\
            & $w$ & Tree expansion factor \\
        \midrule
        \multirow{2}{*}{SVDD\cite{li_svdd_2024}}
            & $B$ & Similar to population size $P$ \\
            & $n_\text{duplicates}$ & Beam width \\
        \bottomrule
    \end{tabular}
    \label{tab:appendix:algorithm_param_descript}
\end{table*}

%% file: tables/method_configuration_table.tex
\begin{table*}[!thb]
    \centering
    \setlength{\tabcolsep}{12pt}
    \caption{
        \textbf{\cref{tab:eval_cross_reward,tab:opi_eval} Hyperparameter Configurations.} 
        These configurations closely match the running time of our best method, \cosyne{} for each alignment objective.
        HPSv2 is slower to evaluate than all other alignment objectives, partially due to its large CLIP  backbone.
        We note that batch size $B$ is interpreted differently by each method, per~\cref{sec:appendix:es_init:others}.
    } 
    \begin{tabular}{lll}
        \toprule
        Algorithm(s) & Objective & Configuration \\
        \midrule
        FKS\cite{singhal_fk_steering_2025} & ImageReward, CLIP, JPEG & $\text{PC}=128$, $t_{\text{start}}=5$, $t_{\text{end}}=45$, $t_{\text{freq}}=1$ \\
        SVDD\cite{li_svdd_2024} & ImageReward, CLIP, JPEG & $B=6$, $n_\text{duplicates}=20$ \\
        DSearch-R\cite{li2025dynamic} & ImageReward, CLIP, JPEG & $B=8$, $n_\text{duplicates}=5$, $w=5$, $r=0.125$ \\
        \midrule
        FKS & HPSv2 & $\text{PC}=16$, $t_{\text{start}}=5$, $t_{\text{end}}=45$, $t_{\text{freq}}=1$ \\
        SVDD & HPSv2 & $B=6$, $n_\text{duplicates}=10$ \\
        DSearch-R & HPSv2 & $B=8$, $n_\text{duplicates}=4$, $w=4$, $r=0.125$\\
        \midrule
        \cosyne{}, SNES, PGPE & ImageReward, CLIP, HPSv2, JPEG & $B=16$, $P=16$ \\
        \bottomrule
    \end{tabular}
    \label{tab:appendix:algorithm_configs}
\end{table*}

%% file: tables/open_img_model_ablation.tex
\begin{table*}[!tb]
\centering
\caption{
\textbf{Open Images Preferences Cross-Model Evaluation} 
The \textit{best‐sample} reward and the \textit{median} reward of the best population achieved by each algorithm on Open Image Preferences across three models: StableDiffusion 1.5, StableDiffusion 3, and PixArt-$\alpha$.
Algorithms are configured in the same manner as in ~\cref{tab:eval_cross_reward}.
Higher rewards are better.
}
\label{tab:eval_cross_model}
\setlength{\tabcolsep}{5pt}      %
\begin{tabular}{l rr rr rr rr rr rr}
\toprule
\multirow{4}{*}{Algorithm}
  & \multicolumn{4}{c}{SD1.5}
  & \multicolumn{4}{c}{SD3}
  & \multicolumn{4}{c}{PixArt-$\alpha$ (LCM)} \\
\cmidrule(lr){2-5}\cmidrule(lr){6-9}\cmidrule(lr){10-13}
  & \multicolumn{2}{c}{CLIP}
  & \multicolumn{2}{c}{ImgReward}
  & \multicolumn{2}{c}{CLIP}
  & \multicolumn{2}{c}{ImgReward}
  & \multicolumn{2}{c}{CLIP}
  & \multicolumn{2}{c}{ImgReward} \\
\cmidrule(lr){2-3}\cmidrule(lr){4-5}
\cmidrule(lr){6-7}\cmidrule(lr){8-9}
\cmidrule(lr){10-11}\cmidrule(lr){12-13}
  & Best & Med. & Best & Med.
  & Best & Med. & Best & Med.
  & Best & Med. & Best & Med. \\
\midrule

Best-of-N & $39.1$\nsp & $34.9$\nsp & $1.49$\nsp & $0.59$\nsp & $39.0$\nsp & $36.0$\nsp & $1.72$\nsp & $1.38$\nsp & $38.8$\nsp & $35.9$\nsp & $1.67$\nsp & $1.25$\nsp \\
Zero–Order   & $39.5$\nsp & $36.4$\nsp & $1.54$\nsp & $0.98$\nsp & $39.4$\nsp & $37.4$\nsp & $1.75$\nsp & $1.60$\nsp & $38.9$\ns & $36.1$\nsp & $1.67$\ns & $1.29$\nsp \\
\midrule
\multicolumn{13}{c}{\textit{Ours: Direct Noise Search}} \\
\midrule
CoSyNE       & $\mathbf{40.7}$\nsp & $\mathbf{38.3}$\nsp & $\mathbf{1.69}$\nsp & $\mathbf{1.47}$\nsp & $\mathbf{40.6}$\nsp & $\mathbf{39.0}$\nsp & $\mathbf{1.82}$\nsp & $\mathbf{1.75}$\nsp & $\mathbf{39.2}$\nsp & $\mathbf{36.8}$\nsp & $\mathbf{1.69}$\nsp & $\mathbf{1.45}$\nsp \\
SNES         & $40.1$\nsp & $37.6$\nsp & $1.57$\nsp & $1.25$\nsp & $39.9$\nsp & $38.1$\nsp & $1.77$\nsp & $1.68$\nsp & $38.9$\ns & $35.8$\nsp & $1.68$\ns & $1.21$\nsp \\
PGPE         & $39.0$\ns & $36.1$\nsp & $1.37$\nsp & $0.88$\nsp & $38.4$\nsp & $36.4$\nsp & $1.67$\nsp & $1.50$\nsp & $38.7$\ns & $35.8$\ns & $1.65$\ns & $1.23$\ns \\

\bottomrule
\end{tabular}
\end{table*}

%% file: tables/dno_long_horizon.tex
\begin{figure*}[!tbp]
    \centering
    \begin{subfigure}[t]{0.32\textwidth}
        \centering
        \includegraphics[width=\linewidth]{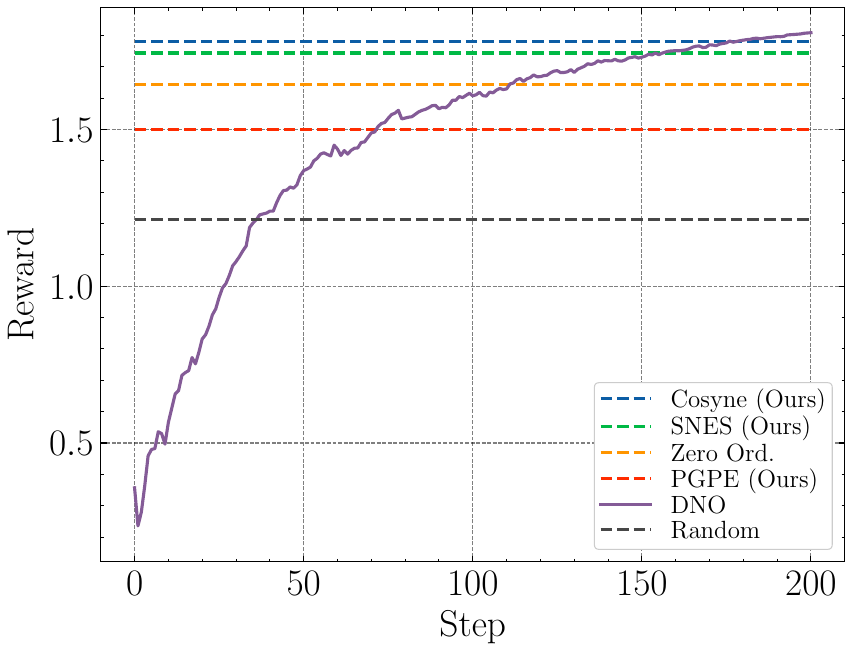}
        \caption{Best-sample reward per step. Horizontal lines show the max reward in \textbf{50} steps.}
        \label{fig:appendix:max_dno_opt}
    \end{subfigure}
    \hfill
    \begin{subfigure}[t]{0.32\textwidth}
        \centering
        \includegraphics[width=\linewidth]{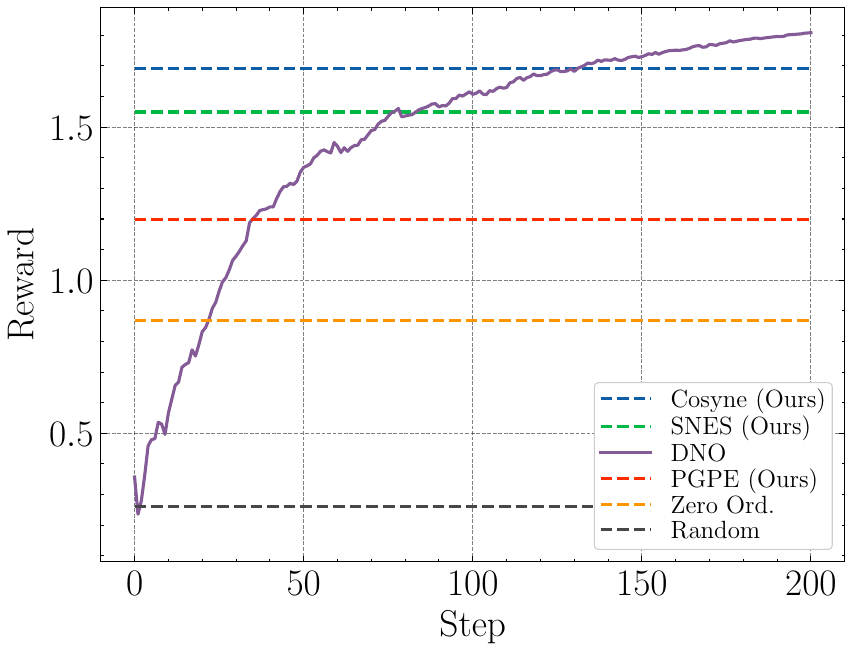}
        \caption{Mean reward per step. Horizontal lines show the max mean in \textbf{50} steps.}
        \label{fig:appendix:mean_dno_opt}
    \end{subfigure}
    \hfill
    \begin{subfigure}[t]{0.32\textwidth}
        \centering
        \includegraphics[width=\linewidth]{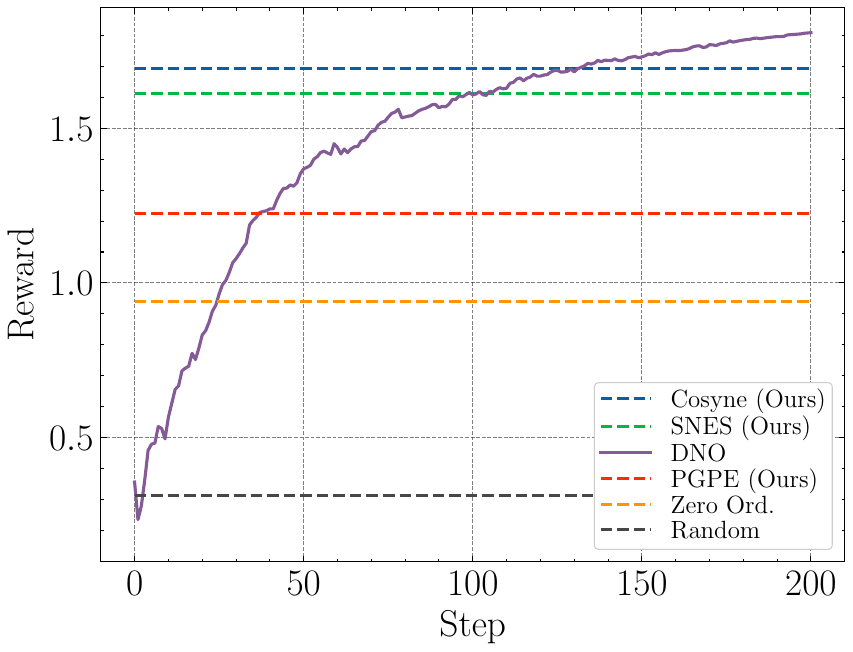}
        \caption{Median reward per step. Horizontal lines show the max median in \textbf{50} steps.}
        \label{fig:appendix:median_dno_opt}
    \end{subfigure}

    \caption{
    \textbf{DNO with Long Optimization Horizon.} ImageReward per step on Open Image Preferences.
    }
    \label{fig:appendix:dno_opt_row}
\end{figure*}

%% file: tables/popsize_ablation.tex
\begin{figure}[!tbp]
    \centering
    \begin{subfigure}[b]{0.48
    \textwidth}
        \includegraphics[width=\textwidth]{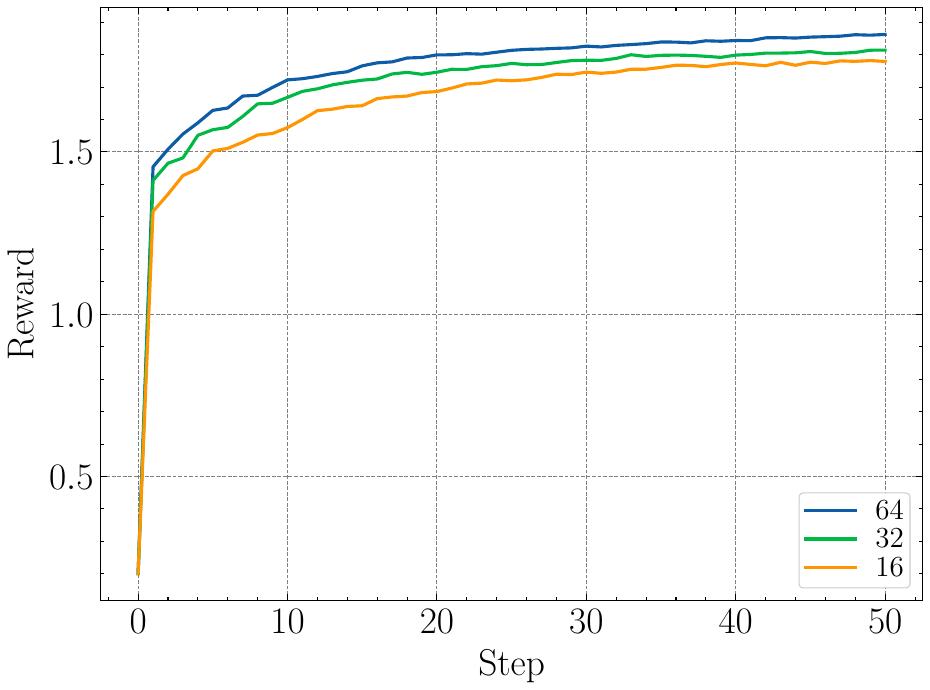}
        \caption{Best-sample reward per step.}
        \label{fig:appendix:cosyne_best}
    \end{subfigure}
    \hfill
    \begin{subfigure}[b]{0.48\textwidth}
        \includegraphics[width=\textwidth]{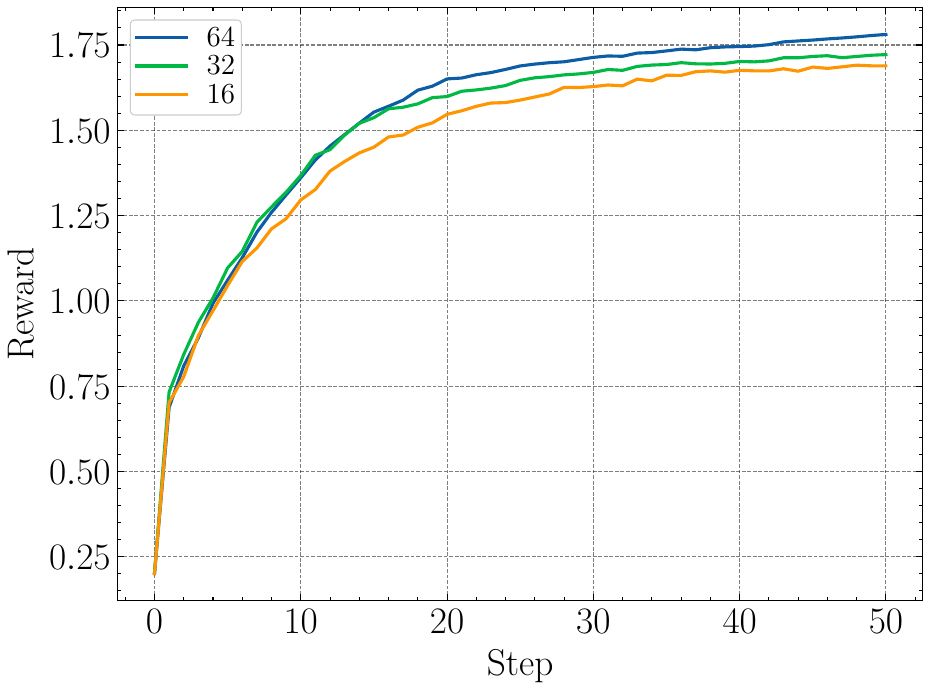}
        \caption{Mean reward per step.}
        \label{fig:appendix:cosyne_mean}
    \end{subfigure}
    \hfill
    \begin{subfigure}[b]{0.48\textwidth}
        \includegraphics[width=\textwidth]{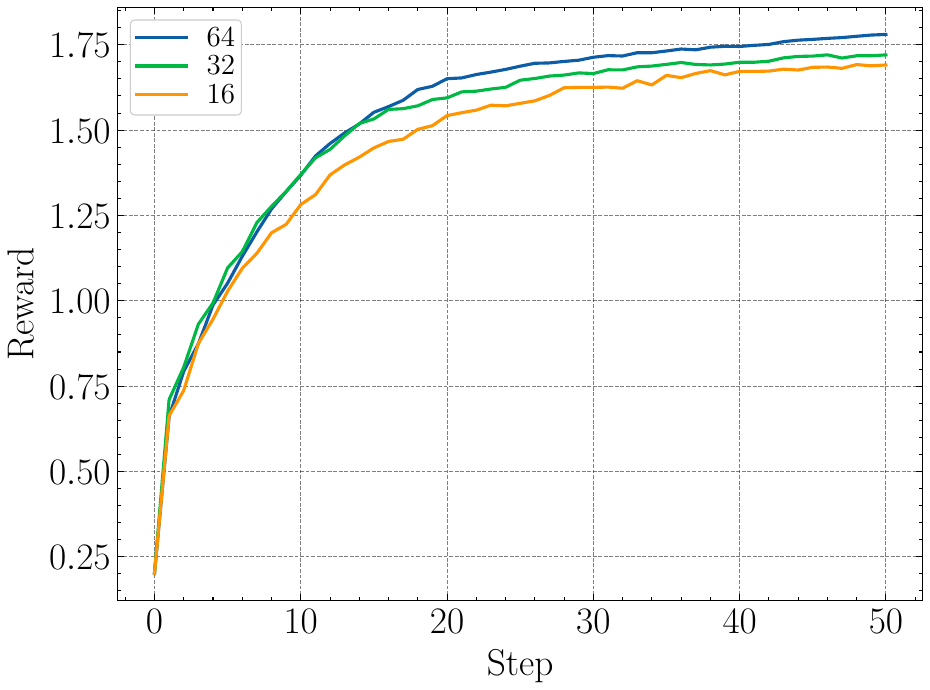}
        \caption{Median reward per step.}
        \label{fig:appendix:cosyne_median}
    \end{subfigure}
    \caption{\textbf{Genetic Algorithm: \cosyne{}} Reward statistics for the \cosyne{} algorithm across optimization step.}
    \label{fig:appendix:cosyne_rewards}
\end{figure}%

\begin{figure}[!tbp]
    \centering
    \begin{subfigure}[b]{0.48
    \textwidth}
        \includegraphics[width=\textwidth]{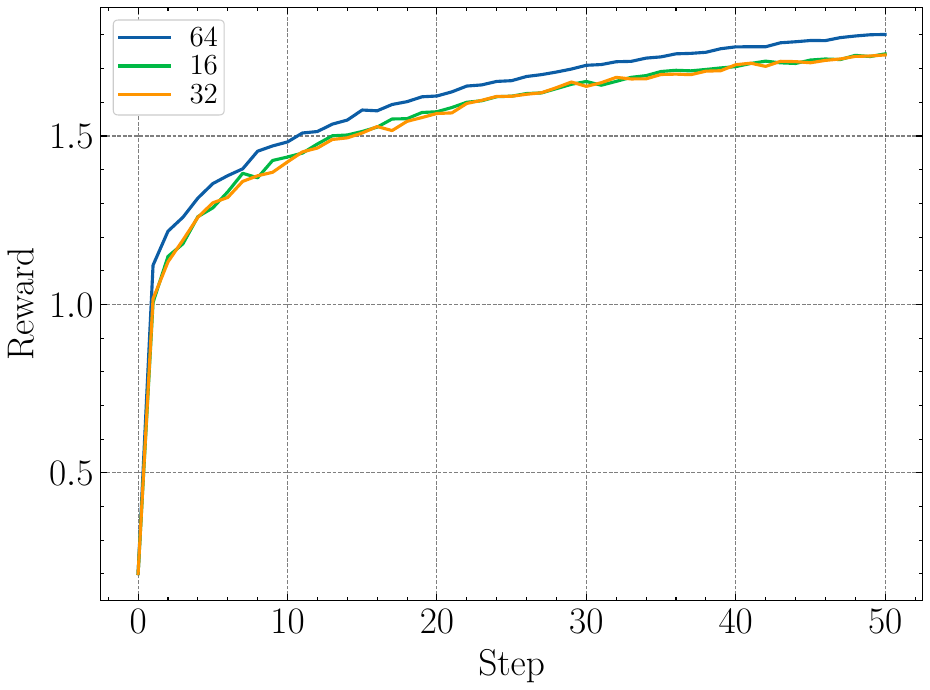}
        \caption{Best-sample reward per step.}
        \label{fig:appendix:snes_best}
    \end{subfigure}
    \hfill
    \begin{subfigure}[b]{0.48\textwidth}
        \includegraphics[width=\textwidth]{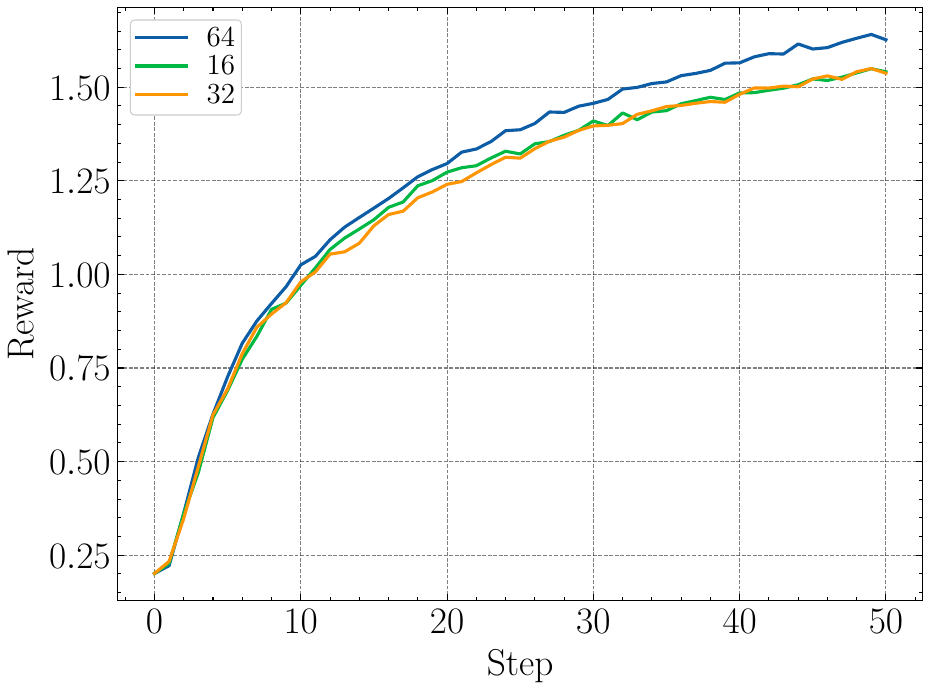}
        \caption{Mean reward per step.}
        \label{fig:appendix:snes_mean}
    \end{subfigure}
    \hfill
    \begin{subfigure}[b]{0.48\textwidth}
        \includegraphics[width=\textwidth]{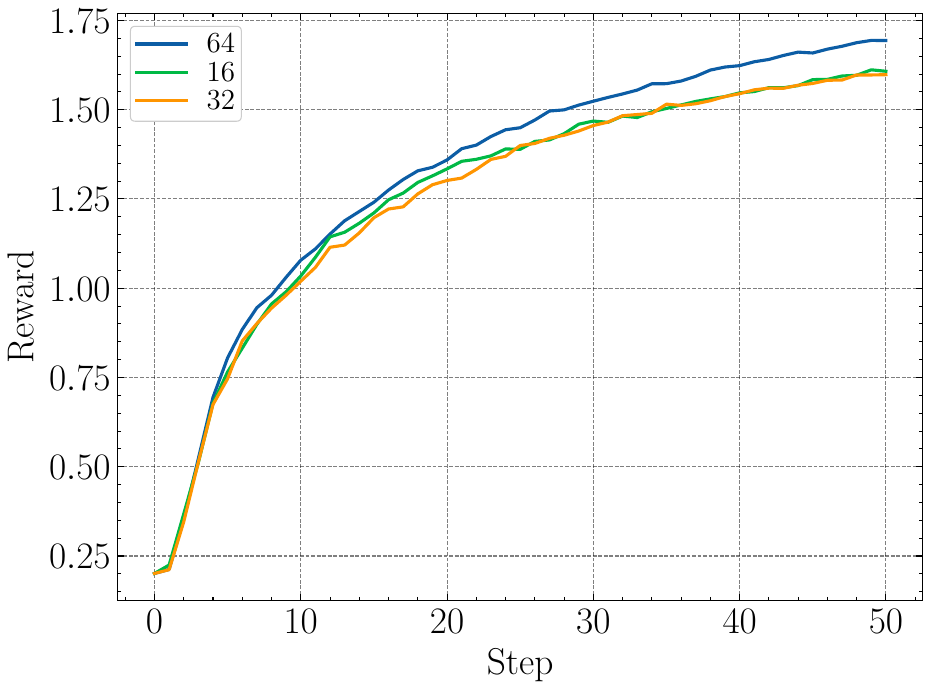}
        \caption{Median reward per step.}
        \label{fig:appendix:snes_median}
    \end{subfigure}
    \caption{\textbf{Evolutionary Strategy: SNES} Reward statistics for the SNES algorithm across optimization step.}
    \label{fig:appendix:snes_rewards}
\end{figure}%

\begin{figure}[!tbp]
    \centering
    \begin{subfigure}[b]{0.48
    \textwidth}
        \includegraphics[width=\textwidth]{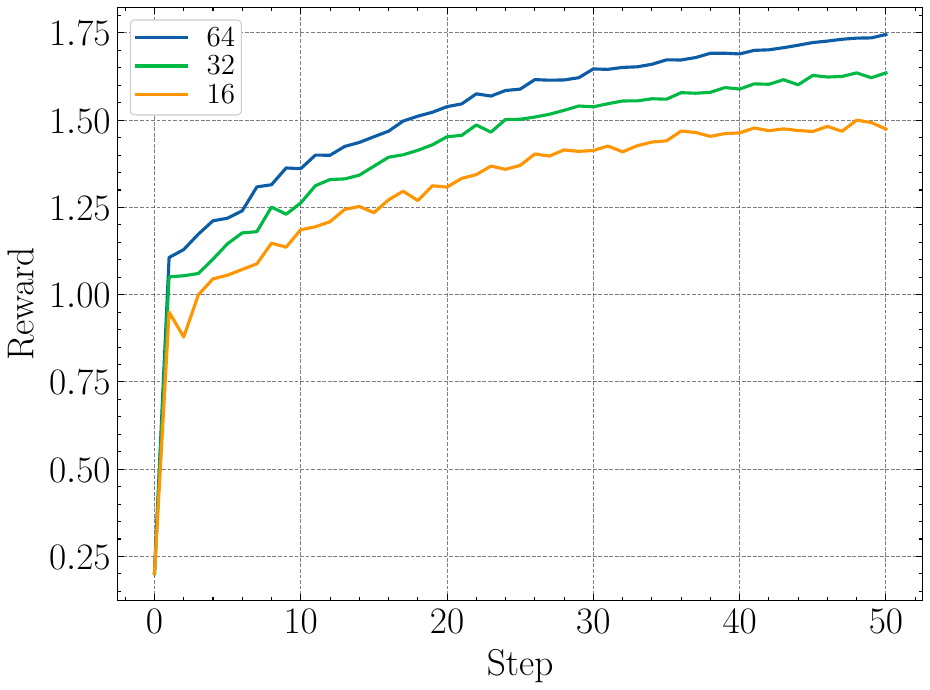}
        \caption{Best-sample reward per step.}
        \label{fig:appendix:pgpe_best}
    \end{subfigure}
    \hfill
    \begin{subfigure}[b]{0.48\textwidth}
        \includegraphics[width=\textwidth]{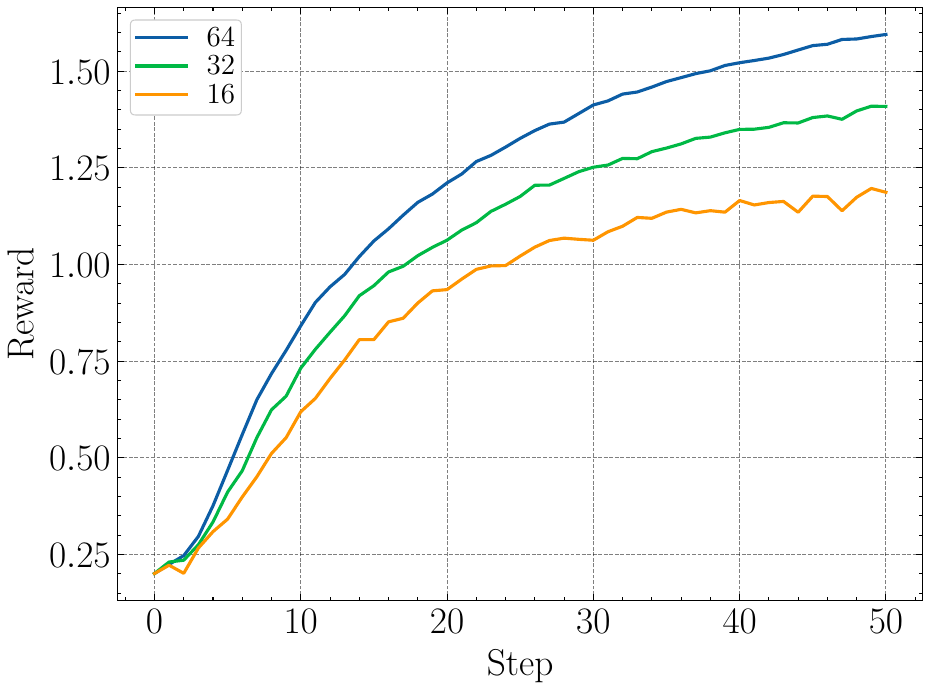}
        \caption{Mean reward per step.}
        \label{fig:appendix:pgpe_mean}
    \end{subfigure}
    \hfill
    \begin{subfigure}[b]{0.48\textwidth}
        \includegraphics[width=\textwidth]{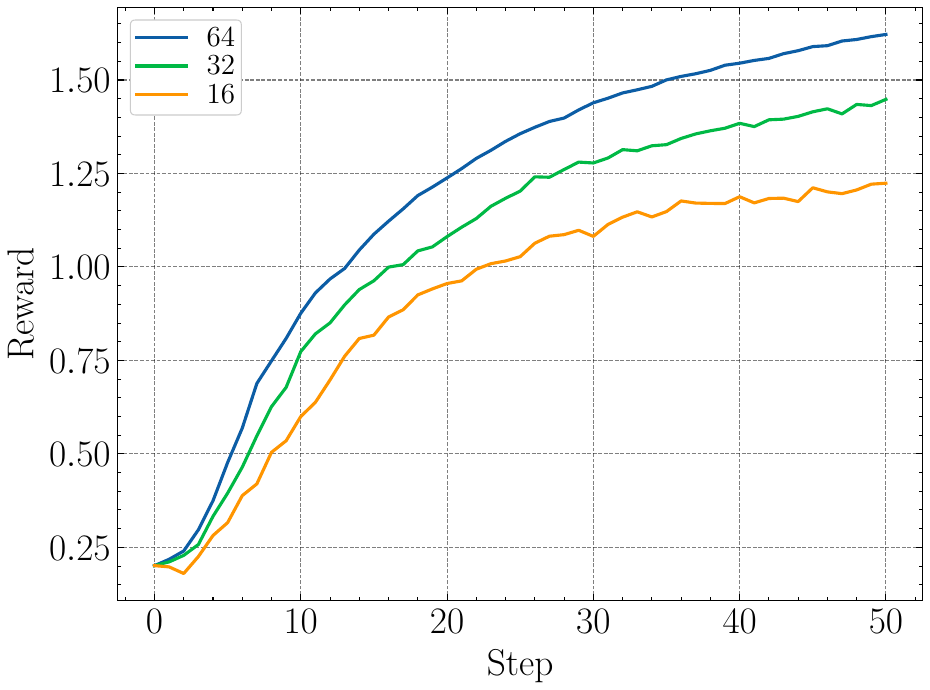}
        \caption{Median reward per step.}
        \label{fig:appendix:pgpe_median}
    \end{subfigure}
    \caption{\textbf{Evolutionary Strategy: PGPE} Reward statistics for the PGPE algorithm across optimization step.}    
    \label{fig:appendix:pgpe_rewards}
\end{figure}

%% file: tables/ga_selection_pressure.tex
\begin{figure}[!tb]
    \centering
    \begin{subfigure}[b]{0.48
    \textwidth}
        \includegraphics[width=\textwidth]{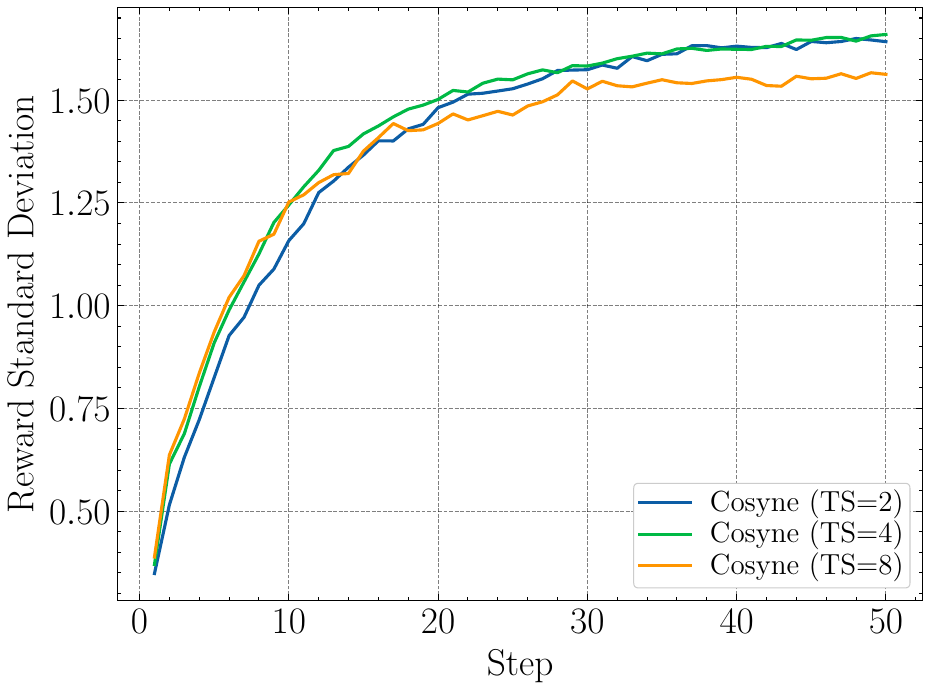}
        \caption{Median reward per step.}
        \label{fig:appendix:selection_median}
    \end{subfigure}
    \hfill
    \begin{subfigure}[b]{0.48\textwidth}
        \includegraphics[width=\textwidth]{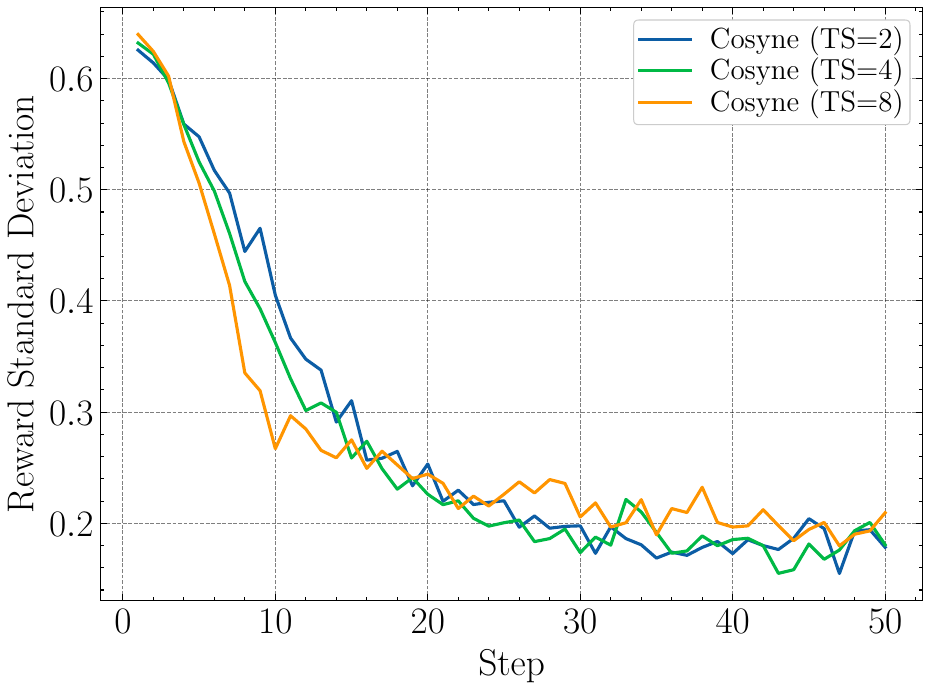}
        \caption{Reward standard deviation per step.}
        \label{fig:appendix:selection_std}
    \end{subfigure}
    \caption{
    \textbf{Genetic Algorithm: \cosyne{}} 
    Reward statistics per
    step, as measured on Open Image Preferences.
    Reward statistics are measured with increasing selection pressure, \ie increasing tournament sizes of 2, 4, and 8. We fix the population size to 16.}    
    \label{fig:appendix:selection}
\end{figure}

%% file: tables/qualitative_diversity.tex
\begin{figure*}[!tbp]
    \centering
    \includegraphics[width=0.95\linewidth]{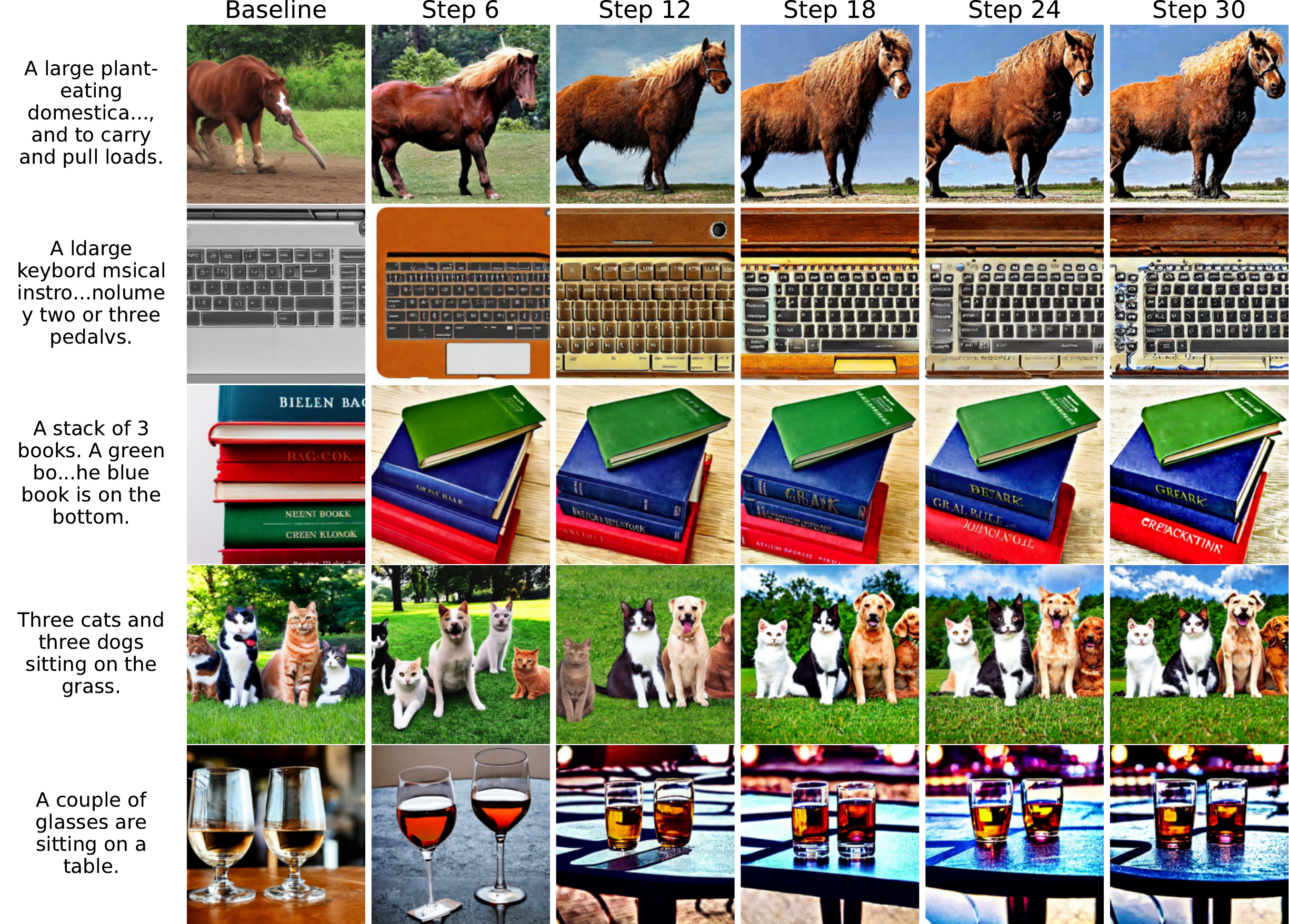}
    \caption{
        \textbf{Qualitative Sample Diversity:} 
        Randomly selected DrawBench prompts evaluated on StableDiffusion-1.5 with ImageReward.
        \cosyne{} was used to perform alignment.
        Across optimization steps, the diversity between samples quickly diminishes.
        Genetic algorithms such as \cosyne{} are particularly vulnerable to this phenomenon.
    }
    \label{fig:appendix:sd_cosyne_img_db}
\end{figure*}

\begin{figure*}[!tbp]
    \centering
    \includegraphics[width=0.92\linewidth]{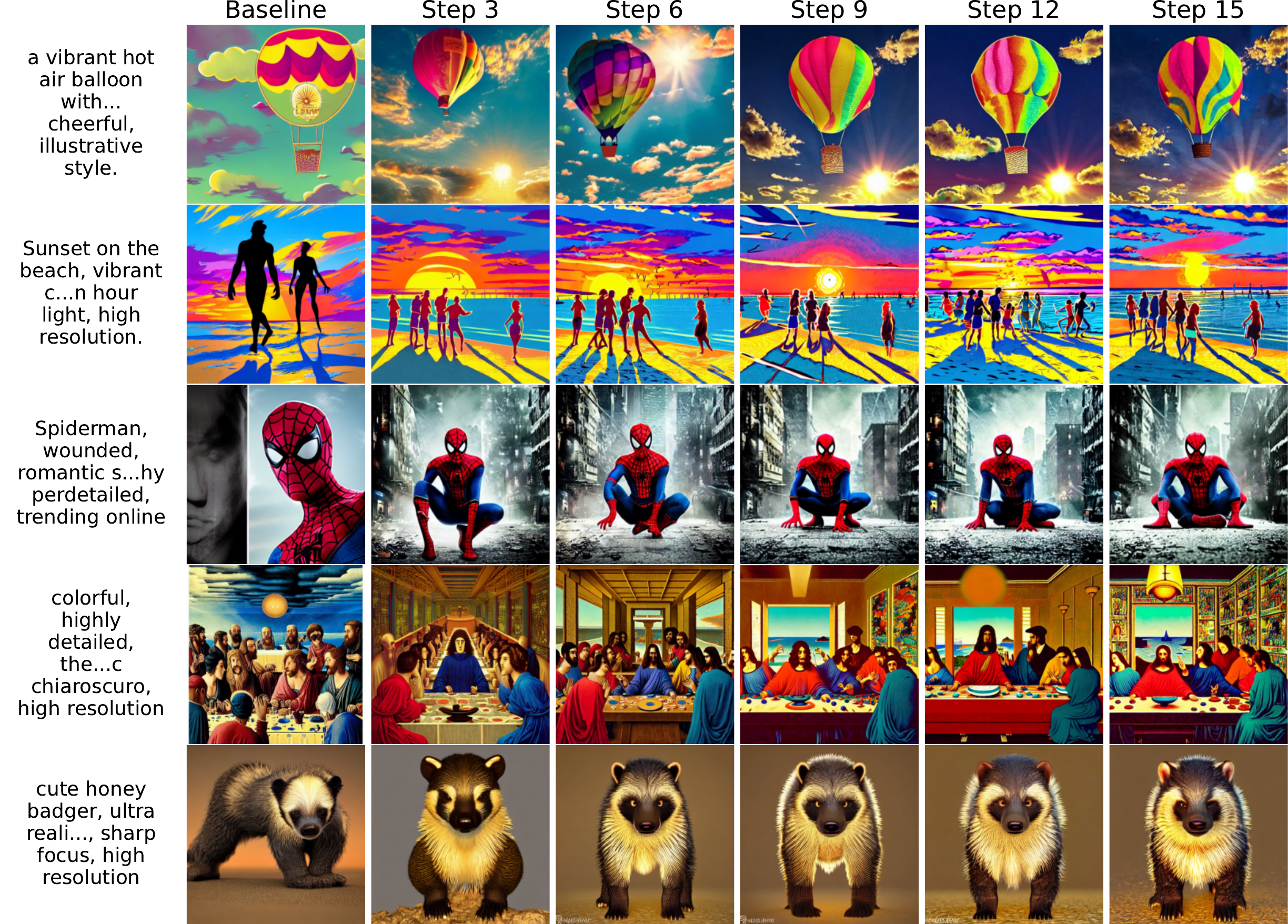}
    \caption{
        \textbf{Qualitative Sample Diversity: }
        Randomly selected Open Image Preference prompts evaluated on StableDiffusion-1.5 with ImageReward.
        \cosyne{} was used to perform alignment.
        Note the low sample diversity across optimization steps.
        We identify this as a consistent occurrence with genetic algorithms such as CoSyne.
    }
    \label{fig:appendix:sd_cosyne_img_open}
\end{figure*}

\begin{figure*}[!tbp]
    \centering
    \includegraphics[width=0.92\linewidth]{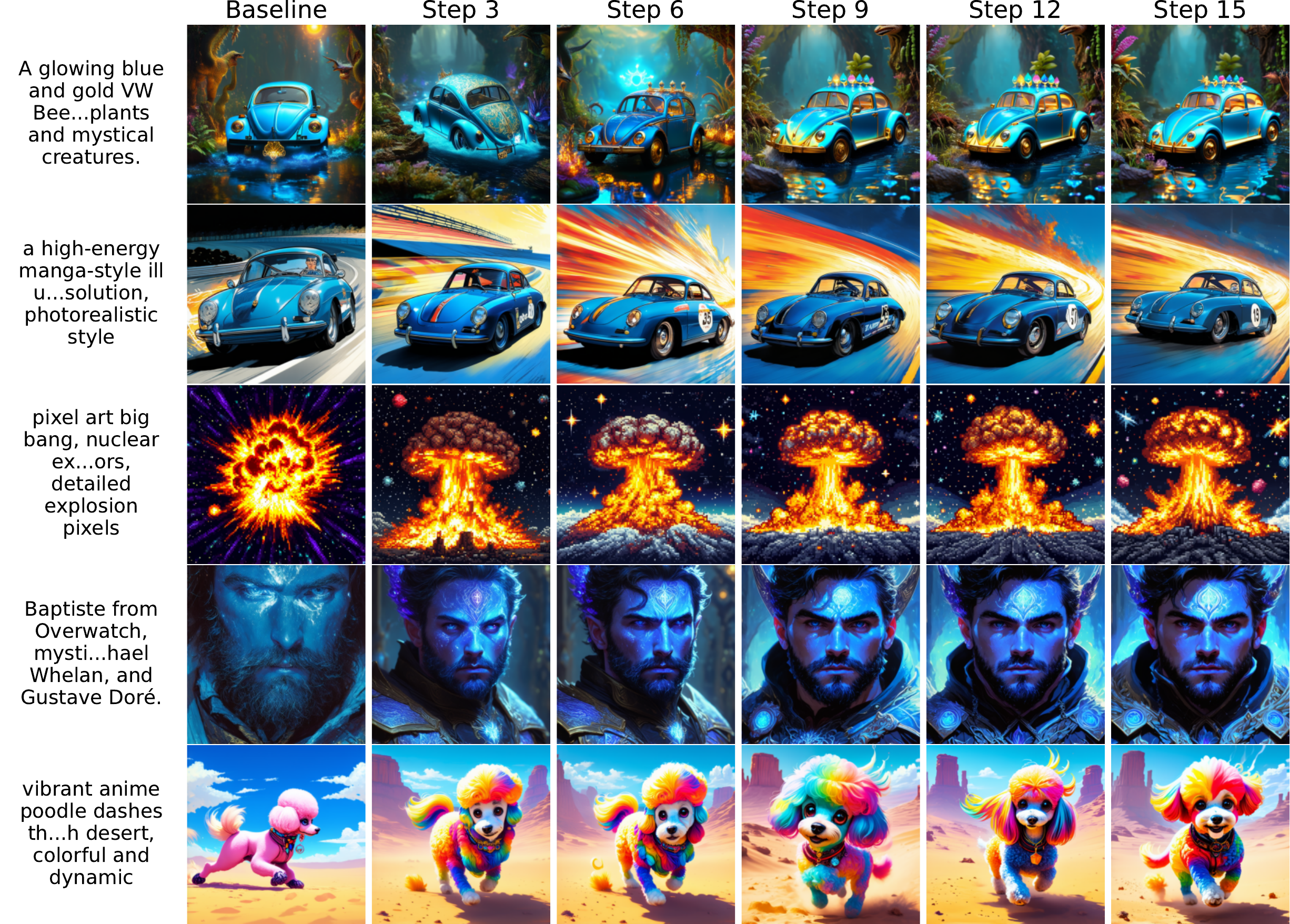}
    \caption{
        \textbf{Qualitative Sample Diversity: }
        Randomly selected Open Image Preference prompts evaluated on StableDiffusion-3 with ImageReward.
        \cosyne{} was used to perform alignment.
        This low sample diversity result on StableDiffusion-3 was also noticed on StableDiffusion-1.5.
    }
    \label{fig:appendix:sd3_cosyne_img_open}
\end{figure*}

\begin{figure*}[!tbp]
    \centering
    \includegraphics[width=0.95\linewidth]{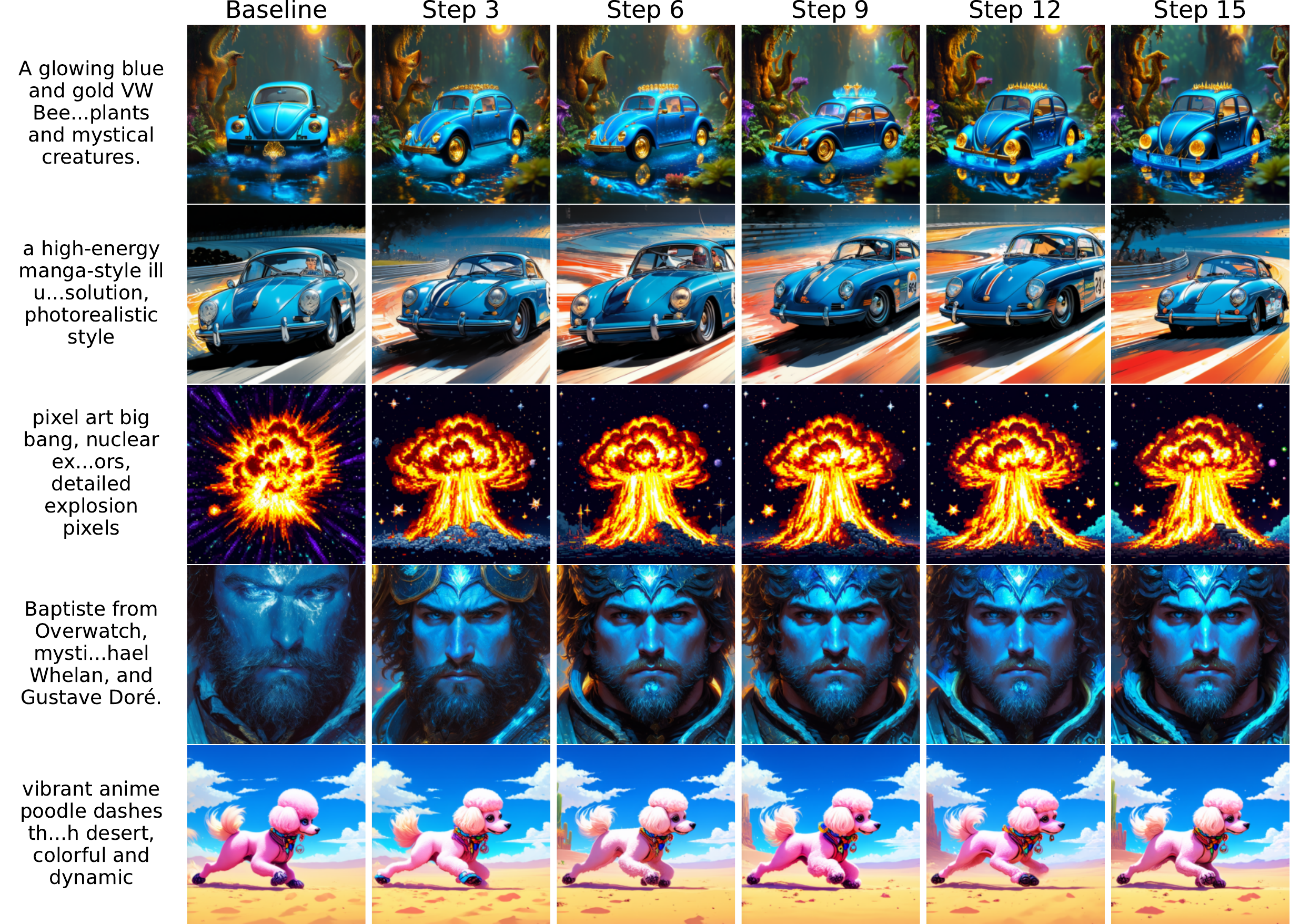}
    \caption{
        \textbf{ Qualitative Sample Diversity: }
        Randomly selected Open Image Preference prompts evaluated on StableDiffusion-3 with ImageReward.
        SNES was used to perform alignment.
    }
    \label{fig:appendix:sd3_snes_img_open}
\end{figure*}

%% file: tables/method_reward_evaluation_counts.tex
\begin{table}[!htb]
    \centering
    \setlength{\tabcolsep}{16pt}
    \caption{
        Number of reward function evaluations per-prompt performed by each method from \cref{tab:eval_cross_reward,tab:opi_eval}.
        If a reward evaluation was batched ($B>1$), we count that as $B$ evaluations.
    }
    \begin{tabular}{lr}
        \toprule
        Algorithm & Reward Evaluations \\
        \midrule
        Best-of-N & 240 \\
        Zero-Order & 240\\
        \midrule 
        FKS & 5258 \\
        SVDD & 5880 \\
        DSearch-R & 7880 \\
        \midrule
        \cosyne{} & 360 \\
        \bottomrule
    \end{tabular}
    \vspace{-1em}
    \label{tab:appendix:reward_evals}
\end{table}

%% file: tables/sd_drawbench_extended.tex
\begin{table*}[!tb]
\centering
\caption{
Extended DrawBench results with ImageReward.
}
\label{tab:appendix:imagereward_extended}
\begin{subtable}[t]{\textwidth}
\centering
\caption{Population-Best Reward}
\begin{tabular}{llrrrr}
\toprule
\textbf{Algorithm} & \textbf{Soln. Space} & \textbf{Mean $\pm$ Std} & \textbf{Median} & \textbf{Min} & \textbf{Max} \\
\midrule
Random             & -             & $1.407 \pm 0.519$ & 1.583 & -0.569 & 1.988 \\
ZeroOrder          & -             & $1.455 \pm 0.530$ & 1.673 & -0.408 & 2.002 \\
DNO                & -             & $0.707 \pm 0.926$ & 0.910 & -1.744 & 1.950 \\
\midrule
Cosyne             & Noise         & $1.614 \pm 0.417$ & 1.765 & -0.093 & 2.002 \\
Cosyne             & Rotation      & $1.532 \pm 0.437$ & 1.659 & -0.259 & 2.001 \\
\midrule
SNES               & Noise         & $1.392 \pm 0.523$ & 1.535 & -0.402 & 1.977 \\
SNES               & Rotation      & $1.338 \pm 0.561$ & 1.482 & -0.744 & 1.983 \\
\midrule
PGPE               & Noise         & $1.261 \pm 0.646$ & 1.419 & -1.166 & 1.978 \\
PGPE               & Rotation      & $1.276 \pm 0.577$ & 1.426 & -0.641 & 1.981 \\
\bottomrule
\end{tabular}
\end{subtable}

\begin{subtable}[t]{\textwidth}
\centering
\caption{Median Reward}
\begin{tabular}{llrrrr}
\toprule
\textbf{Algorithm} & \textbf{Soln. Space} & \textbf{Mean $\pm$ Std} & \textbf{Median} & \textbf{Min} & \textbf{Max} \\
\midrule
Random             & -             & $0.383 \pm 0.844$ & 0.437 & -1.717 & 1.877 \\
ZeroOrder          & -             & $0.966 \pm 0.733$ & 1.093 & -1.532 & 1.976 \\
DNO                & -             & $0.707 \pm 0.926$ & 0.910 & -1.744 & 1.950 \\
\midrule
Cosyne             & Noise         & $1.379 \pm 0.573$ & 1.565 & -0.608 & 1.993 \\
Cosyne             & Rotation      & $1.225 \pm 0.633$ & 1.420 & -0.811 & 1.982 \\
\midrule
SNES               & Noise         & $0.788 \pm 0.801$ & 0.934 & -1.387 & 1.944 \\
SNES               & Rotation      & $0.942 \pm 0.727$ & 1.132 & -1.541 & 1.954 \\
\midrule
PGPE               & Noise         & $0.921 \pm 0.801$ & 1.112 & -1.681 & 1.951 \\
PGPE               & Rotation      & $0.882 \pm 0.731$ & 1.003 & -1.393 & 1.970 \\
\bottomrule
\end{tabular}
\end{subtable}
\end{table*}

\begin{table*}[!tb]
\centering
\caption{
Extended DrawBench results with HPSv2.
}
\label{tab:appendix:hpsv2_extended}
\begin{subtable}[t]{\textwidth}
\centering
\caption{Population-Best Reward}
\begin{tabular}{llrrrr}
\toprule
\textbf{Algorithm} & \textbf{Soln. Space} & \textbf{Mean $\pm$ Std} & \textbf{Median} & \textbf{Min} & \textbf{Max} \\
\midrule
Random             & -             & $0.301 \pm 0.016$ & 0.304 & 0.258 & 0.344 \\
ZeroOrder          & -             & $0.304 \pm 0.017$ & 0.306 & 0.260 & 0.347 \\
DNO                & -             & $0.286 \pm 0.017$ & 0.286 & 0.241 & 0.324 \\
\midrule
Cosyne             & Noise         & $0.310 \pm 0.017$ & 0.311 & 0.271 & 0.352 \\
Cosyne             & Rotation      & $0.307 \pm 0.017$ & 0.308 & 0.267 & 0.347 \\
\midrule
SNES               & Noise         & $0.300 \pm 0.016$ & 0.301 & 0.260 & 0.346 \\
SNES               & Rotation      & $0.300 \pm 0.017$ & 0.302 & 0.250 & 0.342 \\
\midrule
PGPE               & Noise         & $0.299 \pm 0.017$ & 0.301 & 0.256 & 0.342 \\
PGPE               & Rotation      & $0.300 \pm 0.017$ & 0.301 & 0.251 & 0.338 \\
\bottomrule
\end{tabular}
\end{subtable}

\begin{subtable}[t]{\textwidth}
\centering
\caption{Median Reward}
\begin{tabular}{llrrrr}
\toprule
\textbf{Algorithm} & \textbf{Soln. Space} & \textbf{Mean $\pm$ Std} & \textbf{Median} & \textbf{Min} & \textbf{Max} \\
\midrule
Random             & -             & $0.282 \pm 0.016$ & 0.284 & 0.238 & 0.318 \\
ZeroOrder          & -             & $0.290 \pm 0.017$ & 0.292 & 0.245 & 0.336 \\
DNO                & -             & $0.286 \pm 0.017$ & 0.286 & 0.241 & 0.324 \\
\midrule
Cosyne             & Noise         & $0.300 \pm 0.016$ & 0.302 & 0.259 & 0.340 \\
Cosyne             & Rotation      & $0.299 \pm 0.017$ & 0.300 & 0.262 & 0.340 \\
\midrule
SNES               & Noise         & $0.286 \pm 0.016$ & 0.288 & 0.247 & 0.324 \\
SNES               & Rotation      & $0.290 \pm 0.017$ & 0.293 & 0.243 & 0.333 \\
\midrule
PGPE               & Noise         & $0.290 \pm 0.017$ & 0.291 & 0.252 & 0.335 \\
PGPE               & Rotation      & $0.290 \pm 0.017$ & 0.294 & 0.243 & 0.329 \\
\bottomrule
\end{tabular}
\end{subtable}
\end{table*}

\begin{table*}[!tb]
\centering
\caption{
Extended DrawBench results with CLIP.
}
\label{tab:appendix:clip_extended}
\begin{subtable}[t]{\textwidth}
\centering
\caption{Population-Best Reward}
\begin{tabular}{llrrrr}
\toprule
\textbf{Algorithm} & \textbf{Soln. Space} & \textbf{Mean $\pm$ Std} & \textbf{Median} & \textbf{Min} & \textbf{Max} \\
\midrule
Random             & -             & $37.139 \pm 3.665$ & 37.016 & 29.094 & 49.562 \\
ZeroOrder          & -             & $37.590 \pm 3.864$ & 37.719 & 28.000 & 48.344 \\
DNO                & -             & $26.237 \pm 3.638$ & 26.309 & 18.047 & 35.562 \\
\midrule
Cosyne             & Noise         & $38.791 \pm 3.812$ & 38.672 & 30.328 & 50.062 \\
Cosyne             & Rotation      & $38.405 \pm 3.795$ & 37.938 & 30.125 & 50.188 \\
\midrule
SNES               & Noise         & $38.061 \pm 3.751$ & 37.875 & 29.094 & 49.562 \\
SNES               & Rotation      & $37.407 \pm 3.831$ & 37.188 & 29.250 & 48.156 \\
\midrule
PGPE               & Noise         & $37.052 \pm 3.607$ & 36.656 & 29.250 & 47.844 \\
PGPE               & Rotation      & $36.884 \pm 3.763$ & 36.609 & 28.562 & 48.875 \\
\bottomrule
\end{tabular}
\end{subtable}

\begin{subtable}[t]{\textwidth}
\centering
\caption{Median Reward}
\begin{tabular}{llrrrr}
\toprule
\textbf{Algorithm} & \textbf{Soln. Space} & \textbf{Mean $\pm$ Std} & \textbf{Median} & \textbf{Min} & \textbf{Max} \\
\midrule
Random             & -             & $32.867 \pm 3.502$ & 32.812 & 20.562 & 41.750 \\
ZeroOrder          & -             & $34.545 \pm 3.698$ & 34.438 & 24.578 & 44.938 \\
DNO                & -             & $26.237 \pm 3.638$ & 26.309 & 18.047 & 35.562 \\
\midrule
Cosyne             & Noise         & $36.455 \pm 3.614$ & 36.141 & 28.219 & 46.938 \\
Cosyne             & Rotation      & $36.393 \pm 3.712$ & 36.266 & 26.266 & 47.250 \\
\midrule
SNES               & Noise         & $35.623 \pm 3.563$ & 35.328 & 24.547 & 44.969 \\
SNES               & Rotation      & $34.699 \pm 3.662$ & 34.438 & 26.594 & 45.812 \\
\midrule
PGPE               & Noise         & $34.397 \pm 3.455$ & 34.031 & 25.859 & 43.469 \\
PGPE               & Rotation      & $34.772 \pm 3.602$ & 34.516 & 27.172 & 44.750 \\
\bottomrule
\end{tabular}
\end{subtable}
\end{table*}

\begin{table*}[!tb]
\centering
\caption{
Extended DrawBench results with JPEG File Size. Entries are file sizes listed in kilobytes (kB).
}
\label{tab:appendix:jpeg_extended}
\begin{subtable}[t]{\textwidth}
\centering
\caption{Population-Best Reward}
\begin{tabular}{llrrrr}
\toprule
\textbf{Algorithm} & \textbf{Soln. Space} & \textbf{Mean $\pm$ Std} & \textbf{Median} & \textbf{Min} & \textbf{Max} \\
\midrule
Random             & -             & $66.92 \pm 17.82$ & 64.23 & 26.55 & 136.58 \\
ZeroOrder          & -             & $53.35 \pm 18.42$ & 51.47 & 15.89 & 116.90 \\
\midrule
Cosyne             & Noise         & $39.52 \pm 15.88$ & 36.79 & 13.78 & 79.34 \\
Cosyne             & Rotation      & $38.73 \pm 16.25$ & 36.82 & 9.96 & 108.24 \\
\midrule
SNES               & Noise         & $71.35 \pm 21.36$ & 67.54 & 28.47 & 157.80 \\
SNES               & Rotation      & $57.78 \pm 22.40$ & 54.63 & 11.54 & 174.64 \\
\midrule
PGPE               & Noise         & $88.29 \pm 23.68$ & 84.50 & 31.79 & 160.11 \\
PGPE               & Rotation      & $18.29 \pm 12.37$ & 14.76 & 5.00 & 91.34 \\
\bottomrule
\end{tabular}
\end{subtable}

\begin{subtable}[t]{\textwidth}
\centering
\caption{Median Reward}
\begin{tabular}{llrrrr}
\toprule
\textbf{Algorithm} & \textbf{Soln. Space} & \textbf{Mean $\pm$ Std} & \textbf{Median} & \textbf{Min} & \textbf{Max} \\
\midrule
Random             & -             & $105.43 \pm 20.66$ & 103.54 & 60.79 & 203.58 \\
ZeroOrder          & -             & $62.76 \pm 20.65$ & 61.21 & 18.32 & 158.85 \\
DNO                & -             & $94.64 \pm 22.78$ & 91.78 & 47.11 & 174.45 \\
\midrule
Cosyne             & Noise         & $44.76 \pm 16.40$ & 42.42 & 15.42 & 87.25 \\
Cosyne             & Rotation      & $42.28 \pm 16.63$ & 40.86 & 11.91 & 111.33 \\
\midrule
SNES               & Noise         & $77.98 \pm 22.33$ & 73.70 & 32.36 & 168.50 \\
SNES               & Rotation      & $66.10 \pm 23.36$ & 62.25 & 16.72 & 185.45 \\
\midrule
PGPE               & Noise         & $97.02 \pm 24.02$ & 93.35 & 53.09 & 173.79 \\
PGPE               & Rotation      & $20.18 \pm 13.10$ & 16.60 & 5.38 & 94.78 \\
\bottomrule
\end{tabular}
\end{subtable}
\end{table*}

%% file: bib/references.bib
@misc{gaussian_annulus_lecture_2017,
  author       = {Simon Auch and Wolfgang Mulzer},
  title        = {High-Dimensional Space},
  year         = {2017},
  howpublished = {Lecture notes for \textit{Seminar über Algorithmen}, Freie Universität Berlin},
  note         = {Proseminar Theoretische Informatik, 05.11.2017},
  url          = {https://www.inf.fu-berlin.de/lehre/WS17/ALP3/}
}

@misc{li_svdd_2024,
      title={Derivative-Free Guidance in Continuous and Discrete Diffusion Models with Soft Value-Based Decoding}, 
      author={Xiner Li and Yulai Zhao and Chenyu Wang and Gabriele Scalia and Gokcen Eraslan and Surag Nair and Tommaso Biancalani and Shuiwang Ji and Aviv Regev and Sergey Levine and Masatoshi Uehara},
      year={2024},
      eprint={2408.08252},
      archivePrefix={arXiv},
      primaryClass={cs.LG},
      url={https://arxiv.org/abs/2408.08252}, 
}

@misc{singhal_fk_steering_2025,
      title={A General Framework for Inference-time Scaling and Steering of Diffusion Models}, 
      author={Raghav Singhal and Zachary Horvitz and Ryan Teehan and Mengye Ren and Zhou Yu and Kathleen McKeown and Rajesh Ranganath},
      year={2025},
      eprint={2501.06848},
      archivePrefix={arXiv},
      primaryClass={cs.LG},
      url={https://arxiv.org/abs/2501.06848}, 
}

@article{clip2021,
  title={Learning Transferable Visual Models From Natural Language Supervision}, 
  author={Alec Radford and Jong Wook Kim and Chris Hallacy and Aditya Ramesh and Gabriel Goh and Sandhini Agarwal and Girish Sastry and Amanda Askell and Pamela Mishkin and Jack Clark and Gretchen Krueger and Ilya Sutskever},
  year={2021},
  journal={arXiv preprint arXiv:2103.00020}
}

@inproceedings{liu2024latent,
  title={Latent Diffusion Model for Audio: Generation, Quality Enhancement, and Neural Audio Codec},
  author={Liu, Haohe and Wang, Wenwu and Plumbley, Mark D},
  booktitle={Audio Imagination: NeurIPS 2024 Workshop AI-Driven Speech, Music, and Sound Generation},
  year={2024}
}

@article{kong2020diffwave,
  title={Diffwave: A versatile diffusion model for audio synthesis},
  author={Kong, Zhifeng and Ping, Wei and Huang, Jiaji and Zhao, Kexin and Catanzaro, Bryan},
  journal={arXiv preprint arXiv:2009.09761},
  year={2020}
}

@article{ho2022video_diffuion,
  title={Video diffusion models},
  author={Ho, Jonathan and Salimans, Tim and Gritsenko, Alexey and Chan, William and Norouzi, Mohammad and Fleet, David J},
  journal={Advances in Neural Information Processing Systems},
  volume={35},
  pages={8633--8646},
  year={2022}
}

@article{liu2024alignmentdiffusionmodelsfundamentals,
    title={Alignment of Diffusion Models: Fundamentals, Challenges, and Future}, 
    author={Buhua Liu and Shitong Shao and Bao Li and Lichen Bai and Zhiqiang Xu and Haoyi Xiong and James Kwok and Sumi Helal and Zeke Xie},
    journal={arXiv preprint arxiv:2409.07253},
    year={2024}
}

@book{kramer2017genetic,
  title={Genetic Algorithm Essentials},
  author={Kramer, Oliver},
  year={2017},
  publisher={Springer Berlin},
  series={Studies in Computational Intelligence},
  volume={679}
}

@article{ma2025inference_time_scaling,
  title={Inference-time scaling for diffusion models beyond scaling denoising steps},
  author={Ma, Nanye and Tong, Shangyuan and Jia, Haolin and Hu, Hexiang and Su, Yu-Chuan and Zhang, Mingda and Yang, Xuan and Li, Yandong and Jaakkola, Tommi and Jia, Xuhui and others},
  journal={arXiv preprint arXiv:2501.09732},
  year={2025}
}

@article{tang2024inference_dno,
  title={Inference-Time Alignment of Diffusion Models with Direct Noise Optimization},
  author={Tang, Zhiwei and Peng, Jiangweizhi and Tang, Jiasheng and Hong, Mingyi and Wang, Fan and Chang, Tsung-Hui},
  journal={arXiv preprint arXiv:2405.18881},
  year={2024}
}

@article{gomez2008accelerated,
  title={Accelerated Neural Evolution through Cooperatively Coevolved Synapses.},
  author={Gomez, Faustino and Schmidhuber, J{\"u}rgen and Miikkulainen, Risto and Mitchell, Melanie},
  journal={Journal of Machine Learning Research},
  volume={9},
  number={5},
  year={2008}
}

@article{xu2023imagereward,
  title={Imagereward: Learning and evaluating human preferences for text-to-image generation},
  author={Xu, Jiazheng and Liu, Xiao and Wu, Yuchen and Tong, Yuxuan and Li, Qinkai and Ding, Ming and Tang, Jie and Dong, Yuxiao},
  journal={Advances in Neural Information Processing Systems},
  volume={36},
  pages={15903--15935},
  year={2023}
}

@inproceedings{hpswu2023human,
  title={Human preference score: Better aligning text-to-image models with human preference},
  author={Wu, Xiaoshi and Sun, Keqiang and Zhu, Feng and Zhao, Rui and Li, Hongsheng},
  booktitle={Proceedings of the IEEE/CVF International Conference on Computer Vision},
  pages={2096--2105},
  year={2023}
}

@article{hpsv2wu2023human,
  title={Human preference score v2: A solid benchmark for evaluating human preferences of text-to-image synthesis},
  author={Wu, Xiaoshi and Hao, Yiming and Sun, Keqiang and Chen, Yixiong and Zhu, Feng and Zhao, Rui and Li, Hongsheng},
  journal={arXiv preprint arXiv:2306.09341},
  year={2023}
}

@article{kirstain2023pick,
  title={Pick-a-pic: An open dataset of user preferences for text-to-image generation},
  author={Kirstain, Yuval and Polyak, Adam and Singer, Uriel and Matiana, Shahbuland and Penna, Joe and Levy, Omer},
  journal={Advances in Neural Information Processing Systems},
  volume={36},
  pages={36652--36663},
  year={2023}
}

@misc{laionAestheticDataset,
  author = {Schuhmann, Christoph},
  title = {LAION-Aesthetics Dataset},
  howpublished = {\url{https://github.com/LAION-AI/laion-datasets/blob/main/laion-aesthetic.md}},
  year = {2022},
  note = {Accessed: 2024-04-07}
}

@article{dai2023emu,
  title={Emu: Enhancing image generation models using photogenic needles in a haystack},
  author={Dai, Xiaoliang and Hou, Ji and Ma, Chih-Yao and Tsai, Sam and Wang, Jialiang and Wang, Rui and Zhang, Peizhao and Vandenhende, Simon and Wang, Xiaofang and Dubey, Abhimanyu and others},
  journal={arXiv preprint arXiv:2309.15807},
  year={2023}
}

@article{lee2023aligning,
  title={Aligning text-to-image models using human feedback},
  author={Lee, Kimin and Liu, Hao and Ryu, Moonkyung and Watkins, Olivia and Du, Yuqing and Boutilier, Craig and Abbeel, Pieter and Ghavamzadeh, Mohammad and Gu, Shixiang Shane},
  journal={arXiv preprint arXiv:2302.12192},
  year={2023}
}

@article{black2023training,
  title={Training diffusion models with reinforcement learning},
  author={Black, Kevin and Janner, Michael and Du, Yilun and Kostrikov, Ilya and Levine, Sergey},
  journal={arXiv preprint arXiv:2305.13301},
  year={2023}
}

@inproceedings{wallace2024diffusion,
  title={Diffusion model alignment using direct preference optimization},
  author={Wallace, Bram and Dang, Meihua and Rafailov, Rafael and Zhou, Linqi and Lou, Aaron and Purushwalkam, Senthil and Ermon, Stefano and Xiong, Caiming and Joty, Shafiq and Naik, Nikhil},
  booktitle={Proceedings of the IEEE/CVF Conference on Computer Vision and Pattern Recognition},
  pages={8228--8238},
  year={2024}
}

@article{prabhudesai2023aligning,
    title={Aligning text-to-image diffusion models with reward backpropagation},
    author={Prabhudesai, Mihir and Goyal, Anirudh and Pathak, Deepak and Fragkiadaki, Katerina},
    journal={arXiv preprint arXiv:2310.03739},
    year={2023}
}

@article{clark2023directly,
  title={Directly fine-tuning diffusion models on differentiable rewards},
  author={Clark, Kevin and Vicol, Paul and Swersky, Kevin and Fleet, David J},
  journal={arXiv preprint arXiv:2309.17400},
  year={2023}
}

@article{hao2023optimizing,
  title={Optimizing prompts for text-to-image generation},
  author={Hao, Yaru and Chi, Zewen and Dong, Li and Wei, Furu},
  journal={Advances in Neural Information Processing Systems},
  volume={36},
  pages={66923--66939},
  year={2023}
}

@article{feng2022training,
  title={Training-free structured diffusion guidance for compositional text-to-image synthesis},
  author={Feng, Weixi and He, Xuehai and Fu, Tsu-Jui and Jampani, Varun and Akula, Arjun and Narayana, Pradyumna and Basu, Sugato and Wang, Xin Eric and Wang, William Yang},
  journal={arXiv preprint arXiv:2212.05032},
  year={2022}
}

@article{gal2022image,
  title={An image is worth one word: Personalizing text-to-image generation using textual inversion},
  author={Gal, Rinon and Alaluf, Yuval and Atzmon, Yuval and Patashnik, Or and Bermano, Amit H and Chechik, Gal and Cohen-Or, Daniel},
  journal={arXiv preprint arXiv:2208.01618},
  year={2022}
}

@article{manas2024improving,
  title={Improving text-to-image consistency via automatic prompt optimization},
  author={Ma{\~n}as, Oscar and Astolfi, Pietro and Hall, Melissa and Ross, Candace and Urbanek, Jack and Williams, Adina and Agrawal, Aishwarya and Romero-Soriano, Adriana and Drozdzal, Michal},
  journal={arXiv preprint arXiv:2403.17804},
  year={2024}
}

@inproceedings{wallace2023end,
  title={End-to-end diffusion latent optimization improves classifier guidance},
  author={Wallace, Bram and Gokul, Akash and Ermon, Stefano and Naik, Nikhil},
  booktitle={Proceedings of the IEEE/CVF International Conference on Computer Vision},
  pages={7280--7290},
  year={2023}
}

@article{li2025dynamic,
  title={Dynamic Search for Inference-Time Alignment in Diffusion Models},
  author={Li, Xiner and Uehara, Masatoshi and Su, Xingyu and Scalia, Gabriele and Biancalani, Tommaso and Regev, Aviv and Levine, Sergey and Ji, Shuiwang},
  journal={arXiv preprint arXiv:2503.02039},
  year={2025}
}

@article{zhou2024golden,
  title={Golden noise for diffusion models: A learning framework},
  author={Zhou, Zikai and Shao, Shitong and Bai, Lichen and Xu, Zhiqiang and Han, Bo and Xie, Zeke},
  journal={arXiv preprint arXiv:2411.09502},
  year={2024}
}

@article{uehara2025inference,
  title={Inference-Time Alignment in Diffusion Models with Reward-Guided Generation: Tutorial and Review},
  author={Uehara, Masatoshi and Zhao, Yulai and Wang, Chenyu and Li, Xiner and Regev, Aviv and Levine, Sergey and Biancalani, Tommaso},
  journal={arXiv preprint arXiv:2501.09685},
  year={2025}
}

@article{miao2024t2vsafetybench,
  title={T2vsafetybench: Evaluating the safety of text-to-video generative models},
  author={Miao, Yibo and Zhu, Yifan and Yu, Lijia and Zhu, Jun and Gao, Xiao-Shan and Dong, Yinpeng},
  journal={Advances in Neural Information Processing Systems},
  volume={37},
  pages={63858--63872},
  year={2024}
}

@book{eiben2015introduction,
  title={Introduction to evolutionary computing},
  author={Eiben, Agoston E and Smith, James E},
  year={2015},
  publisher={Springer}
}

@book{simon2013evolutionary,
  title={Evolutionary optimization algorithms},
  author={Simon, Dan},
  year={2013},
  publisher={John Wiley \& Sons}
}

@article{wierstra2014natural,
  title={Natural evolution strategies},
  author={Wierstra, Daan and Schaul, Tom and Glasmachers, Tobias and Sun, Yi and Peters, Jan and Schmidhuber, J{\"u}rgen},
  journal={The Journal of Machine Learning Research},
  volume={15},
  number={1},
  pages={949--980},
  year={2014}
}

@incollection{miikkulainen2024evolving,
  title={Evolving deep neural networks},
  author={Miikkulainen, Risto and Liang, Jason and Meyerson, Elliot and Rawal, Aditya and Fink, Dan and Francon, Olivier and Raju, Bala and Shahrzad, Hormoz and Navruzyan, Arshak and Duffy, Nigel and others},
  booktitle={Artificial intelligence in the age of neural networks and brain computing},
  pages={269--287},
  year={2024},
  publisher={Elsevier}
}

@inproceedings{young2015optimizing,
  title={Optimizing deep learning hyper-parameters through an evolutionary algorithm},
  author={Young, Steven R and Rose, Derek C and Karnowski, Thomas P and Lim, Seung-Hwan and Patton, Robert M},
  booktitle={Workshop on Machine Learning in High-performance Computing Environments},
  pages={1--5},
  year={2015}
}

@article{such2017deep,
  title={Deep neuroevolution: Genetic algorithms are a competitive alternative for training deep neural networks for reinforcement learning},
  author={Such, Felipe Petroski and Madhavan, Vashisht and Conti, Edoardo and Lehman, Joel and Stanley, Kenneth O and Clune, Jeff},
  journal={arXiv preprint arXiv:1712.06567},
  year={2017}
}

@article{salimans2017evolution,
  title={Evolution strategies as a scalable alternative to reinforcement learning},
  author={Salimans, Tim and Ho, Jonathan and Chen, Xi and Sidor, Szymon and Sutskever, Ilya},
  journal={arXiv preprint arXiv:1703.03864},
  year={2017}
}

@inproceedings{lehman2011evolving,
  title={Evolving a diversity of virtual creatures through novelty search and local competition},
  author={Lehman, Joel and Stanley, Kenneth O},
  booktitle={Conference on Genetic and Evolutionary Computation},
  pages={211--218},
  year={2011}
}

@inproceedings{secretan2008picbreeder,
  title={Picbreeder: evolving pictures collaboratively online},
  author={Secretan, Jimmy and Beato, Nicholas and D Ambrosio, David B and Rodriguez, Adelein and Campbell, Adam and Stanley, Kenneth O},
  booktitle={Proceedings of the SIGCHI conference on human factors in computing systems},
  pages={1759--1768},
  year={2008}
}

@inproceedings{tian2022modern,
  title={Modern evolution strategies for creativity: Fitting concrete images and abstract concepts},
  author={Tian, Yingtao and Ha, David},
  booktitle={International conference on computational intelligence in music, sound, art and design (part of evostar)},
  pages={275--291},
  year={2022},
  organization={Springer}
}

@article{kumar2024automating,
  title={Automating the Search for Artificial Life with Foundation Models},
  author={Kumar, Akarsh and Lu, Chris and Kirsch, Louis and Tang, Yujin and Stanley, Kenneth O and Isola, Phillip and Ha, David},
  journal={arXiv preprint arXiv:2412.17799},
  year={2024}
}

@article{toklu2023evotorch,
  title={Evotorch: Scalable evolutionary computation in python},
  author={Toklu, Nihat Engin and Atkinson, Timothy and Micka, Vojt{\v{e}}ch and Liskowski, Pawe{\l} and Srivastava, Rupesh Kumar},
  journal={arXiv preprint arXiv:2302.12600},
  year={2023}
}

@article{drawbench2022,
  title={Photorealistic text-to-image diffusion models with deep language understanding},
  author={Saharia, Chitwan and Chan, William and Saxena, Saurabh and Li, Lala and Whang, Jay and Denton, Emily L and Ghasemipour, Kamyar and Gontijo Lopes, Raphael and Karagol Ayan, Burcu and Salimans, Tim and others},
  journal={Advances in Neural Information Processing Systems},
  volume={35},
  pages={36479--36494},
  year={2022}
}

@inproceedings{rombach2022high,
  title={High-resolution image synthesis with latent diffusion models},
  author={Rombach, Robin and Blattmann, Andreas and Lorenz, Dominik and Esser, Patrick and Ommer, Bj{\"o}rn},
  booktitle={Proceedings of the IEEE/CVF conference on computer vision and pattern recognition},
  pages={10684--10695},
  year={2022}
}

@inproceedings{esser2024scaling,
  title={Scaling rectified flow transformers for high-resolution image synthesis},
  author={Esser, Patrick and Kulal, Sumith and Blattmann, Andreas and Entezari, Rahim and M{\"u}ller, Jonas and Saini, Harry and Levi, Yam and Lorenz, Dominik and Sauer, Axel and Boesel, Frederic and others},
  booktitle={International Conference on Machine Learning},
  year={2024}
}

@article{chen2023pixart,
  title={Pixart-$\alpha$: Fast training of diffusion transformer for photorealistic text-to-image synthesis},
  author={Chen, Junsong and Yu, Jincheng and Ge, Chongjian and Yao, Lewei and Xie, Enze and Wu, Yue and Wang, Zhongdao and Kwok, James and Luo, Ping and Lu, Huchuan and others},
  journal={arXiv preprint arXiv:2310.00426},
  year={2023}
}

@misc{huggingface_image_preferences,
  author       = {Berenstein, David  and Burtenshaw, Ben and Vila, Daniel and van Strien, Daniel and Paul, Sayak and Vi, Ame and Tsaban, Linoy},
  title        = {Open Preference Dataset for Text-to-Image Generation by the HuggingFace Community},
  howpublished = {\url{https://huggingface.co/blog/image-preferences}},
  note         = {Accessed: 2025-04-15},
  year         = {2024}
}

@misc{data_is_better_together_open_image_preferences_v1_flux_dev_lora_2024,
  author       = {{Data Is Better Together}},
  title        = {{open-image-preferences-v1-flux-dev-lora}},
  year         = {2024},
  publisher    = {Hugging Face},
  url          = {https://huggingface.co/data-is-better-together/open-image-preferences-v1-flux-dev-lora},
  note         = {Accessed: 2025-04-17},
}

@article{gu2024aligning,
  title={Aligning target-aware molecule diffusion models with exact energy optimization},
  author={Gu, Siyi and Xu, Minkai and Powers, Alexander and Nie, Weili and Geffner, Tomas and Kreis, Karsten and Leskovec, Jure and Vahdat, Arash and Ermon, Stefano},
  journal={Advances in Neural Information Processing Systems},
  year={2024}
}

@article{wang2024fine,
  title={Fine-tuning discrete diffusion models via reward optimization with applications to dna and protein design},
  author={Wang, Chenyu and Uehara, Masatoshi and He, Yichun and Wang, Amy and Biancalani, Tommaso and Lal, Avantika and Jaakkola, Tommi and Levine, Sergey and Wang, Hanchen and Regev, Aviv},
  journal={arXiv preprint arXiv:2410.13643},
  year={2024}
}

@inproceedings{sohl-dickstein_deep_2015,
    author = {Sohl-Dickstein, Jascha and Weiss, Eric A. and Maheswaranathan, Niru and Ganguli, Surya},
    title = {Deep {Unsupervised} {Learning} using {Nonequilibrium} {Thermodynamics}},
    booktitle={International Conference on Machine Learning},
    year={2015}
}

@inproceedings{fei2023gradient,
  title={Gradient-free textual inversion},
  author={Fei, Zhengcong and Fan, Mingyuan and Huang, Junshi},
  booktitle={Proceedings of the 31st ACM International Conference on Multimedia},
  pages={1364--1373},
  year={2023}
}

@article{miller1995genetic,
  title={Genetic algorithms, tournament selection, and the effects of noise},
  author={Miller, Brad L and Goldberg, David E and others},
  journal={Complex systems},
  volume={9},
  number={3},
  pages={193--212},
  year={1995},
  publisher={[Champaign, IL, USA: Complex Systems Publications, Inc., c1987-}
}
